\documentclass[acmtog]{acmart}
\AtBeginDocument{%
  }

\setcopyright{acmlicensed}
\copyrightyear{2026}
\acmYear{2026}
\acmDOI{XXXXXXX.XXXXXXX}

\acmJournal{TOG}

\usepackage[table]{xcolor}
\usepackage{colortbl}

\newcommand{\kd}[1]{{\color{black}#1}}

\newtheorem{theorem}{Theorem}

\usepackage{booktabs}
\usepackage{multirow}
\usepackage[normalem]{ulem}
\useunder{\uline}{\ul}{}
\usepackage{algorithm}
\usepackage{algorithmic}
\definecolor{algohighlight}{RGB}{220, 244, 237}

\begin{document}

\title{TopSDF: Topological Signed Distance Field for High-Quality Triangle Mesh Extraction from 3D Gaussian Splatting}

\author{Khoi Do}
\orcid{0000-0002-2566-5637}
\affiliation{%
  \institution{Trinity College Dublin}
  \city{Dublin}
  \country{Ireland}}
\email{dokh@tcd.ie}

\author{Yu-Kun Lai}
\email{LaiY4@cardiff.ac.uk}
\affiliation{%
  \institution{Cardiff University}
  \city{Cardiff}
  \state{Wales}
  \country{United Kingdom}
}

\author{Binh-Son Hua}
\affiliation{%
  \institution{Trinity College Dublin}
  \city{Dublin}
  \country{Ireland}}
\email{binhson.hua@tcd.ie}

\begin{abstract}
Recent advances in 3D Gaussian Splatting (3DGS) have demonstrated its effectiveness for photorealistic novel view synthesis. However, extracting clean, watertight, explicit surface meshes from 3DGS remains a critical challenge. Previous methods approximate geometry via discrete Marching Tetrahedra, where each grid vertex is assigned a single scalar signed distance value. At complex geometric boundaries and thin structures, this suboptimal parameterization forces severe compromises across shared vertices, causing gradient collisions and geometric frustration. Furthermore, existing optimization objectives neglect element shape conditioning, producing degenerate triangles. We introduce TopSDF, a framework that models the signed distance state as a multivariate Gaussian random field. By learning latent topological covariances across tetrahedral vertices, our formulation expands local degrees of freedom for SDF estimates, allowing the candidate mesh vertices to be placed flexibly along the bipolar edges in the tetrahedra as driven by such topological correlation. Concurrently, we directly integrate variational meshing energies into the joint optimization, coupled with an energy-driven Gaussian densification strategy to improve triangle mesh quality. Our experiments across Mip-NeRF 360, Tanks \& Temples, and ScanNet++ show that TopSDF achieves state-of-the-art mesh-based novel view rendering (+0.86 dB PSNR on Mip-NeRF 360) and triangle-mesh quality (from 0.37 to 0.49 aspect ratio).
\end{abstract}

\ccsdesc[500]{Computing methodologies~Reconstruction}
\ccsdesc[300]{Computing methodologies~Mesh models}
\ccsdesc[300]{Computing methodologies~Point-based models}
\ccsdesc[100]{Computing methodologies~Parametric curve and surface models}
\ccsdesc[100]{Computing methodologies~Machine learning approaches}

\keywords{3D Gaussian Splatting, Surface Reconstruction, Differentiable Mesh, Mesh-Guided Optimization}


\begin{teaserfigure}
  \centering
  \includegraphics[width=\textwidth]{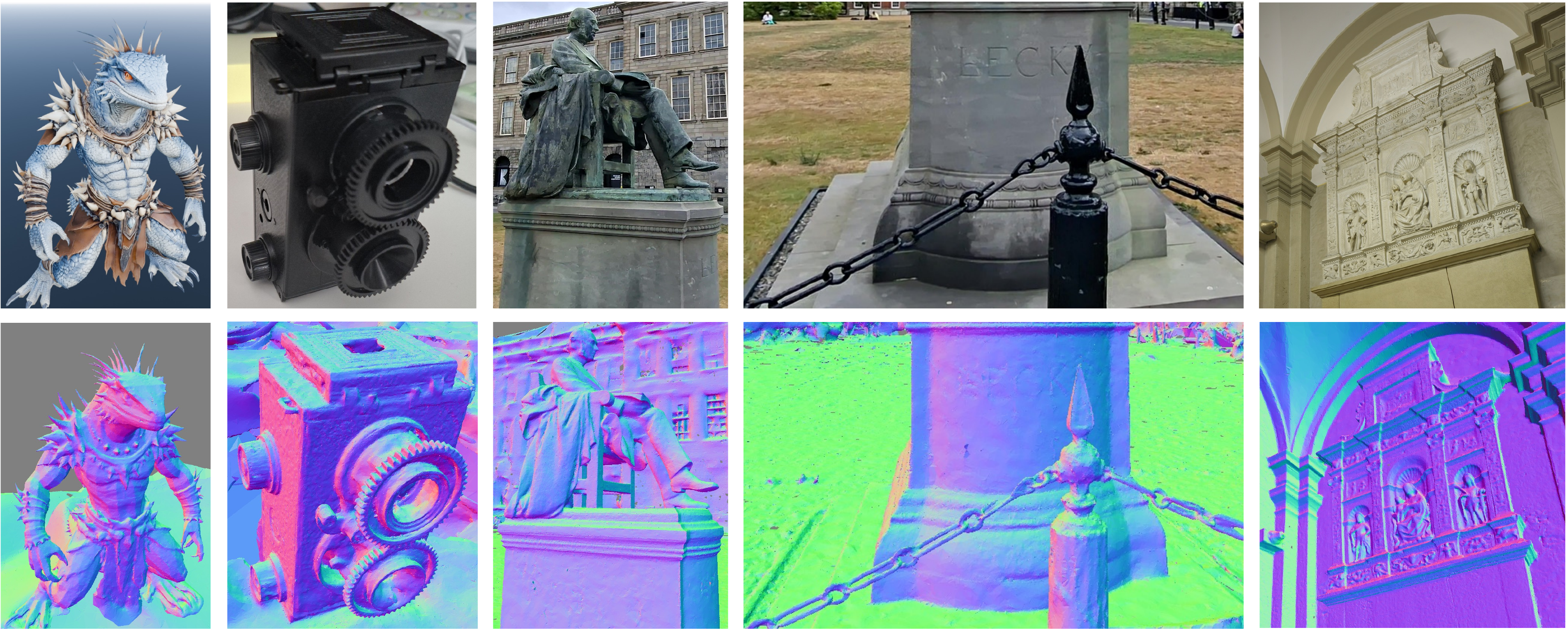}
  \Description{Teaser figure showing explicit surface reconstructions and normal maps.}
  \captionof{figure}{Overview of mesh-based novel view view rendering and triangle mesh extracted by our framework. The top row shows RGB renderings of extracted triangle meshes across diverse object-centric and unbounded scene from real-world captures. The bottom row displays the corresponding extracted surface normal maps, illustrating reconstructed geometric structures, sharp boundary silhouettes, and topological surface continuity.}
  \label{fig:teaser}
\end{teaserfigure}
\maketitle

\section{Introduction}
Reconstructing highly detailed 3D models from multi-view photographs is a long-standing challenge. State-of-the-art methods are based on radiance fields and novel view synthesis (NVS) using NeRF~\cite{nerf} (Neural Radiance Field) and 3D Gaussian Splatting~\cite{3dgs} (3DGS). Primitive-based radiance fields like 3DGS~\cite{3dgs} capture intricate micro-details and render in real-time, but their discrete point-based representation implies the lack of a straightforward pathway for extracting an accurate and high-quality triangle mesh. These methods optimize a radiance field representation by minimizing photometric losses between rendered images and their corresponding ground truth views. To obtain surface geometry, it is necessary to apply mesh extraction as a post-processing or concurrent step~\cite{milo}, which predicts surface location and orientation for mesh conversion from the learned radiance fields. 

Recent methods extract meshes by first approximating signed distance fields (SDFs) using depth and normal rendering~\cite{gw, radegs, gggs} and then obtain triangle meshes using a marching tetrahedra algorithm. 
%
Hybrid models~\cite{milo, gof, sof, 2dgs, come, gggs} additionally use mesh-to-Gaussians alignment losses and appearance or geometry regularization terms. 
However, a fundamental limitation of these methods is their reliance on naive tetrahedral grids and signed distance values at discrete grid vertices, which comes at the expense of flexibility for placing the mesh vertices during marching tetrahedra procedure. Each grid vertex only stores a single distance value~\cite{gw, come, milo}, leading to compromised approximation of the \kd{surface mesh vertices} within a one-ring neighborhood, and hence degradations in complex geometry regions, thin structures, or fine details. 


\kd{We aim to address such compromised approximation through regularising SDFs at grid vertices using topological correlation.
} 
Acknowledging these topological relationships allows us to determine how the underlying surface is influenced by surrounding local grid vertices, from which the placement of the surface mesh vertices could be flexibly enhanced. 
To learn these correlations, we propose to model the SDF in a tetrahedra as a Gaussian random field, where we can \kd{model} the topological correlation using covariance matrices. We construct topological energy as the negative log-likelihood of the SDF state. 
\kd{To optimize this energy, we design a differentiable closed-form solution that can be easily integrated in the 3DGS training loop without introducing additional inner iterative optimization.
The closed-form solution leads to SDF updates at grid vertices that better reflect the underlying surfaces.
}

To derive this closed-form solution, we leverage Reproducing Kernel Hilbert Space~\cite{treves2006topological} (RKHS) to derive an equivalent form of the log-likelihood, and then use the Representer Theorem~\cite{represent_theory} to obtain the minimizer. 
This formulation allows us to estimate SDF updates at an arbitrary 3D point, which has the form of a weighted linear combination of feature correlations among grid vertices' SDF. 
We integrate this closed-form SDF estimates to the training of 3DGS, allowing the joint learning of 3D Gaussian attributes and tetrahedra SDFs for novel view synthesis and triangle mesh extraction. 
We demonstrate that the closed-form solution leads to more accurate reconstruction, state-of-the-art mesh-based novel view synthesis, and improves the convergence rate of the joint optimization of 3D Gaussians and mesh extraction.

Concurrently, another vital yet systematically neglected factor in 3DGS mesh extraction is triangle mesh quality. Existing pipelines yield irregular, sliver-filled surface triangulations that require expensive, lossy post-processing, leading to poor quality of mesh-based novel view synthesis. We propose to leverage variational meshing energies~\cite{variational_tetrahedral_meshing, odt} for regularising tetrahedral element shapes and dihedral angles, resulting in more regular triangle meshes during the joint optimization. 
Additionally, we use such energies to guide the densification during 3DGS training. We perform energy integration at each 3D Gaussian from the corresponding tetrahedra to identify Gaussians that produce elongated triangles and subsequently split these Gaussians, therefore effectively dicing elongated triangles toward more equilateral ones. 


In summary, our core contributions are:
\begin{itemize}
    \item \kd{We propose modeling the SDF as a continuous random variable over a Reproducing Kernel Hilbert Space (RKHS) and derive a closed-form solution for SDF estimates on tetrahedra grid vertices.}
    \item \kd{We prove theoretically that the closed-form solution leads to accurate reconstruction via increasing the degrees of freedom for SDF estimates at grid vertices. We then prove the improved convergence rate of the joint training of 3D Gaussians and mesh extraction.}
    \item \kd{We propose to a new Gaussian densification strategy based on per-tetrahedron meshing energy, improving the triangle quality of the extracted triangle mesh.}
    \item We conduct extensive experiments and demonstrate state-of-the-art performance on mesh-based novel view synthesis (NVS), surface reconstruction, and triangle mesh quality on multiple datasets under diverse settings, along with comprehensive ablation and empirical studies validating our theoretical findings.
\end{itemize}

\section{Related Work}

\paragraph{Novel View Synthesis.} While Neural Radiance Fields~\cite{nerf} (NeRF) pioneered neural rendering using MLP-parameterized ray-marching---later improved in antialiasing~\cite{mipnerf360, mipnerf} and speed~\cite{instantngp, tensorf, plenoxels}---3D Gaussian Splatting~\cite{3dgs, ren2025fastgs} fundamentally revolutionized novel view synthesis by replacing volumetric integration with rasterized Gaussian primitives for real-time performance. Although concurrent extensions progressively optimize 3DGS for scale~\cite{3dgsmarkov, taming3dgs} and image fidelity~\cite{mip-splatting}, these visual-synthesis techniques systematically neglect the challenge of recovering clean, explicit triangle surface meshes. Triangle-based approaches~\cite{trianglesplating, meshsplatting, demtrianglesplatting} enhance rendering quality using an unconstrained triangle soup rather than an explicit manifold mesh. 

\paragraph{Isosurface Extraction.} Isosurfacing methods have a long history that has been built on different domains, including Delaunay Triangulation~\cite{delaunay1934sphere, vism, tetweave}, particle simulation~\cite{witkinsurvey}, and active contours~\cite{cohensurvey}. Reconstruction algorithms based on Marching algorithms have a limitation of sub-optimality, where the reconstruction is highly localized in space, with suboptimal assumptions about the input samples. In this research, we address this sub-optimality problem by relying on the topological feature assigned at each grid vertex, where the candidate vertices are placed more flexibly, enabling the capture of complex surface geometry where needed, without over-resolving ordinary smooth regions.

In the past, some marching methods allowed placing multiple bipolar vertices along a single bipolar edge~\cite{yamazaki_nomanifold, wywill_polygonalization}. However, these approaches do not produce manifold triangle meshes and rely heavily on deterministic algorithms, lacking differentiability. Recently, Subgrid Marching Tetrahedra~\cite{smt} follows this spirit that allows the surface mesh to intersect multiple times along a single edge of a tetrahedron, which not only resolves the complex geometry structure, but also reduces the number of tetrahedra needed for constructing the volume grid. However, the applicability of Subgrid Marching Tetrahedra (SMT) is limited in 3DGS since there is no explicit surface mesh for multiple sampling along tetrahedral edges. Furthermore, to match the results of 3DGS and the extracted mesh, the mesh needs to be differentially rasterized as well, while SMT relies on discrete marching, restricting the differentiability. 

Another branch of research focuses on adaptive grid refinement, which helps resolve the fine features~\cite{pis, mdc, flexicubes}; however, these approaches are difficult to avoid the cracks transitioning between levels of hierarchy (i.e. relying on octree grid), and break the fixed, regular layout needed for high-performance parallel implementation. For example, FlexiCubes proposed a differentiable method that relies on Dual Marching Cubes~\cite{dmc}, producing a non-manifold, self-intersecting triangle mesh. Conversely, our method extracts a manifold mesh from 3DGS, resolving the fine details.

\paragraph{Surface Extractions from 3D Gaussians.} While implicit SDF-based approaches~\cite{li2023neuralangelo, neus, volsdf} produce highly consistent geometries, they require prohibitively long optimization periods. Alternatively, mesh baking techniques~\cite{bog, bakedsdf} rely on a two-stage extraction process that accelerates rendering but compromises true geometric accuracy. Lastly, discrete mesh optimization~\cite{Munkberg_2022_CVPR, flexicubes, tetweave} performs well on single objects but lacks the scalability needed for complex environments. 

Leveraging 3DGS efficiency, recent surface-extraction methods take two main routes. Hybrid approaches co-optimize Gaussians with implicit fields~\cite{neus, gaussianudf, 3dgsr, gsdf, 3dgs-sdf-infer}, but MLP bottlenecks restrict their scalability. Alternatively, direct regularization methods modify the primitives or constraints: 2DGS~\cite{2dgs} adopts lower-dimensional primitives; TopoSurfel~\cite{toposurfel} initializes Gaussians with Planar-based Gaussian Splatting~\cite{pgsr} (PGSR), then conducts post-training with Gaussian surfels; RaDe-GS~\cite{radegs}, PGSR~\cite{pgsr}, Quadratic Gaussian Splatting~\cite{quad3dgs}, and VCR-GauS~\cite{vcr-gaus} enforce multi-view and rendering consistency; while SuGaR~\cite{sugar}, GOF~\cite{gof}, and SOF~\cite{sof} introduce custom extraction pipelines. GGGS~\cite{gggs} and Gaussian Wrapping~\cite{gw}, in contrast, propose depth rasterization based on Object-as-Volumes~\cite{oav}. CoMe~\cite{come}, instead, proposes color and normal invariance loss functions based on VastGaussian~\cite{vastgs}. Crucially, recent approaches that rely on tetrahedral grids (including Gaussian Wrapping~\cite{gw}, CoMe~\cite{come}, and MILo~\cite{milo}) share a fundamental limitation: their single-scalar vertex SDF representation is suboptimal, leading to severe geometric compromises and inaccurate surface extraction. Compared to these methods, our topological signed distance field allows us to place the mesh vertices more flexibly and avoid interpolation compromises while offering improved training convergence. 

\section{Backgrounds \& Motivation}\label{sec:background}

\paragraph{3D Gaussians Splatting.} 3DGS~\cite{3dgs} represents a scene as $N$ anisotropic Gaussian primitives, where the $i$-th 3D Gaussian primitive is characterized by a center (mean) $\mu_i\in\mathbb{R}^{3}$ and a full 3D covariance matrix $\Sigma_i\in\mathbb{R}^{3\times3}$. For a given point $x \in \mathbb{R}^3$, the spatial distribution of the primitive is defined as $G_i(x) = \exp \left( -\frac{1}{2} (x - \mu_i)^T \Sigma_i^{-1} (x - \mu_i) \right)$. 

To ensure positive semi-definiteness during optimization, the covariance matrix is factorized as $\Sigma_i = R_i S_i S_i^T R_i^T$, where $R_i\in\mathbb{R}^{3\times3}$ is a rotation matrix and $S_i\in\mathbb{R}^{3\times3}$ is a diagonal scale matrix. We denote $\alpha_i$ and $c_i$ as the opacity and color of Gaussian $i$. 

\kd{
Recent approaches tackling surface reconstruction from 3DGS rely primarily on the combination of Delaunay Tetrahedralization and Marching Tetrahedra, where a tetrahedral grid is constructed with approximated signed distances at discrete grid vertices, extracting the surface via linear edge interpolation. Particularly, during optimization, deformable pivots are sampled from Gaussians along with the estimated SDF at each pivot. From these pivots, a tetrahedral grid is constructed, which is then used to extract a triangle mesh using a marching tetrahedra algorithm. The extracting process can be a post-processing task~\cite{gw, come, 2dgs} or the intergrated into the training loop to guide the Gaussians~\cite{milo, toposurfel}. Here, we focus on integrating the extraction process into the training loop, enabling both Gaussians and mesh to be guided to align using depth-normal alignment loss. Note that, during optimization, the pivots' positions could be updated based on the learning of the Gaussian primitives. 
}


\paragraph{Delaunay Tetrahedralization.} Given a set of 3D pivots sampled from Gaussian centers $V = \{\mathbf{v}_1, \dots, \mathbf{v}_M\}$, Delaunay Tetrahedralization~\cite{delaunay1934sphere} partitions the convex hull of $V$ into a connected tetrahedral mesh $\mathcal{T}$. 

Formally, for any tetrahedron $t \in \mathcal{T}$ defined by its four non-coplanar vertices $\{\mathbf{v}_i, \mathbf{v}_j, \mathbf{v}_k, \mathbf{v}_l\}$, the unique 3D circumsphere passing through these four vertices must not contain any other point $\mathbf{v} \in V$ strictly within its interior. The obtained tetrahedral mesh $\mathcal{T}$ is commonly utilized as a spatial grid scaffold along with SDFs for surface extraction~\cite{milo, gw, come}. 

\begin{figure*}[!ht]
    \centering
    \includegraphics[width=\linewidth]{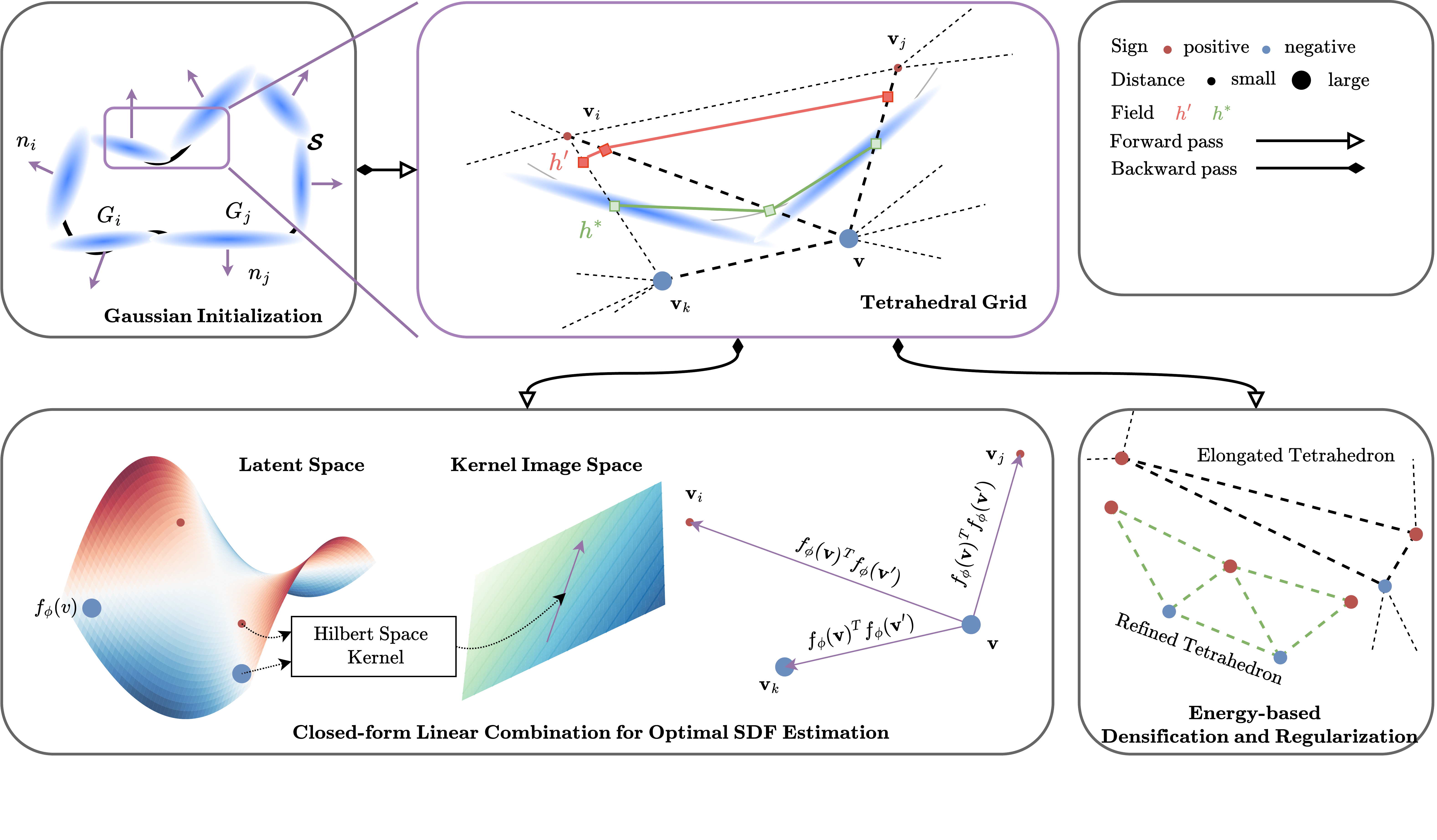}
    \Description{Overview of the framework pipeline and joint training loop.}
    \caption{Framework overview and training loop (formalized in Algorithm~\ref{alg:training_loop}). We revisit the \kd{suboptimality} in triangle mesh extraction from 3DGS caused by the gradient frustration in a single SDF tetrahedral grid. We propose to assign learnable topological features based on Reproducing Kernel Hilbert Space at each grid vertex, allowing us to expand the \kd{SDF} degrees of freedom in localized space. During training, we dynamically adjust candidate vertices by obtaining per-edge relation feature $f_\phi(\mathbf{v})^Tf_\phi(\mathbf{v})$, and then calculate the topological SDF $h^*$ using Eq.~(\ref{eq:inter_opt_sdf}) at each grid vertex. The training procedure is as follows: Starting with normal-oriented Gaussians, we sample a set of pivots, assign learnable normals $\mathbf{u}$ and SDF $\mathbf{y}$, and obtain the topological SDF $h^*$. We then perform Marching Tetrahedra on the Delaunay grid formed by pivots; the extracted mesh is regularized via normal-depth alignment loss, while the Tetrahedral grid is regularized by meshing energies, which is used for Gaussian densification.}
    \label{fig:overview}
\end{figure*}

\paragraph{Marching Tetrahedra.} Marching Tetrahedra~\cite{isosurfaces} (MT) is an iso-surface extraction algorithm designed to reconstruct a polygonal triangle mesh $\mathcal{M}$ from a volumetric scalar field. In our framework, the scalar field $\mathbf{y}$ represents the signed distance values defined over the Delaunay tetrahedral mesh $\mathcal{T}$, guaranteeing the extraction of a manifold triangular mesh.

The algorithm processes the volumetric space one tetrahedron $t \in \mathcal{T}$ at a time, evaluating the sign of the distance values at its four vertices. Any edge connecting two vertices $\mathbf{v}_i, \mathbf{v}_j$ with opposite signs ($\mathbf{y}_i \cdot \mathbf{y}_j < 0$) is identified as a bipolar edge, indicating where the zero-level surface intersects. The exact 3D spatial coordinates of the intersection $\mathbf{v}_c$---which becomes a vertex of the extracted triangle mesh $\mathcal{M}$---are determined via linear interpolation of the spatial coordinates, weighted by their absolute signed distance values:
\begin{align}\label{eq:interpolation}
    \mathbf{v}_c = (1 - \zeta)\mathbf{v}_i + \zeta \mathbf{v}_j,~\textrm{where}~\zeta = \frac{\mathbf{y}_i}{\mathbf{y}_i - \mathbf{y}_j}.
\end{align}

\paragraph{Suboptimality.}

Conventionally, each vertex in the tetrahedra holds only a single scalar signed distance value. Because a single vertex is shared across multiple incident bipolar edges, assigning a single scalar forces a severe mathematical compromise across all connected edges. Consequently, the extracted zero-level surface fails to align with the true scene geometry. \kd{A naive approach is to assign multiple signed distance values at each grid vertex to adjust the per-edge candidate mesh vertex. However, as aforementioned, during optimization, the tetrahedral grid is changed over time, leading to the disappearance of Delaunay edges; thus, deriving a mathematical formulation for the whole grid is a challenging task.} 

To break this interpolation bottleneck, \citet{tetweave} proposes to place a learnable spherical harmonic at each grid vertex, then obtain a rescaled signed distance value, named directional signed distance. We argue that this mechanism does not fully solve the underlying issue, as it considers only a single pair of edges, lacking modeling of the crossing level within a tetrahedron. Otherwise, the proposed formulation lacks a theoretical analysis to derive the solution form, relying on an uncertainty solution.

Recently, \citet{smt} proposed an upgraded version of Marching Tetrahedra, which does not rely on SDF but on the number of intersection points along the bipolar edges. Therefore, the lookup table used in the algorithm is modified to produce a manifold mesh, which preserves fine details and thin structures. However, the grid structure of this algorithm is based on a voxel grid where each voxel is split into tetrahedra, which does not apply to a non-uniform tetrahedral grid, leading to a non-manifold mesh. Previous methods~\cite{gw, come} have shown that a voxel grid is not suitable for mesh extraction from 3DGS since it prevents scalability and resolving the complex geometry. As a result, this modified version does not apply to 3DGS.

\paragraph{Motivation.}
In contrast, we elevate the SDF from a suboptimal discrete parameterization to a continuous multivariate Gaussian random field, allowing us to model the topological connectivities of a one-ring neighborhood given a single tetrahedral vertex. 
This approach allows us to place interpolated vertices along the bipolar edges by utilizing topological correlation from grid vertices.  
This comes from the fact that we can learn the covariance among grid vertices in the Gaussian random field, and then take advantage of these correlations to update SDF at tetrahedral grid vertices more flexibly, which allows us to obtain more accurate surface extraction.
To formalize this approach, we leverage Reproducing Kernel Hilbert Space~\cite{treves2006topological} (RKHS), a common tool in Functional Analysis and Machine Learning, for deriving the topological correlation. We derive a form of topological energy such that we can utilize the Representer Theorem~\cite{represent_theory} to obtain the closed-form solution of SDF, in which the topological feature is learned at each grid vertex, bringing more degrees of freedom to SDFs estimation, thus directly improving surface accuracy and rendering quality in the joint optimization of 3D Gaussians and surface extraction.

\section{Our Method}\label{sec:method}

An overview of our method is in Fig.~\ref{fig:overview}. We start by modeling the state of SDF as a multivariate Gaussian random field, allowing us to learn the topological relationship among grid vertices to flexibly place mesh vertices along the bipolar edges. Rather than relying on a neural network to learn the topological feature, which tends to overfit on small data and requires extensive hyperparameter tuning, we derive the closed-form solution of the SDF, which is simple, plug-and-play, but effective in addressing the sub-optimality of the tetrahedral grid. We show the derivation step by step in detail in Section~\ref{subsec:ltsdf}. 

To obtain high-quality triangle meshes and corresponding mesh-based novel view synthesis, we propose to regularize the mesh extraction with tetrahedral meshing energies. Since poor-quality triangle meshes degrade the quality of rendering and lighting accuracy, we propose using local and global meshing energies on the tetrahedral mesh to improve the shape of the tetrahedra, resulting in a more regular triangle mesh. We detail the mesh energies in  Section~\ref{subsec:mesh-energy}. 

\subsection{Topological SDF via Reproducing Kernel Hilbert Space}\label{subsec:ltsdf}

\paragraph{Signed Distance Field State Modeling.} While assigning a single deterministic signed distance value to each tetrahedral vertex is standard practice in differentiable surface extraction, we identify a fundamental Degrees of Freedom (DoF) bottleneck in this discrete formulation. In Marching Tetrahedra, a single vertex $\mathbf{v}_i$ simultaneously dictates the zero-crossings of all incident bipolar edges in its 1-ring neighborhood. At complex topological transitions---such as sharp geometric corners or thin boundaries ---this suboptimal coupling induces severe geometric frustration: conflicting rendering gradients from distinct edges collide at the shared vertex, trapping the discrete field in a compromised, suboptimal state~\cite{tetweave, smt}. 

To overcome this bottleneck, we model the SDF value $y \in \mathbb{R}$ at any tetrahedral grid vertex $x \in \mathbb{R}^3$ as a random variable, parameterizing the state of the entire tetrahedral grid as a multivariate Gaussian random field $\mathbf{y} \in \mathbb{R}^M$ over $\mathbf{x} \in \mathbb{R}^{M \times 3}$, where $M$ is the number of grid vertices. 
For each tetrahedron $t$, we define $\mathbf{y}_t\in\mathbb{R}^4 = h_{\theta_t}(\mathbf{x}_t)$ as the SDF values at four grid vertices $\mathbf{x}_t\in\mathbb{R}^{4\times 3}$ of the tetrahedron. For simplicity, we use $\mathbf{y}_\theta$ to denote $\mathbf{y}_t$ with learnable parameter $\theta_t$ for SDF estimation from grid vertices $\mathbf{x}_t$ within a single tetrahedron. Analogously, we use $\mathbf{x}$ to denote $\mathbf{x}_t$.

The joint probability distribution over grid vertices of a tetrahedron is formulated as:
\begin{align}
    p(\mathbf{y}_\theta, \mathbf{\mu}, \mathbf{\Sigma}_\phi) = \frac{1}{\sqrt{(2\pi)^M |\mathbf{\Sigma}_\phi|}} \exp\left( -\frac{1}{2} (\mathbf{y}_\theta - \boldsymbol{\mu})^T \mathbf{\Sigma}_{\phi}^{-1} (\mathbf{y}_\theta - \boldsymbol{\mu}) \right),
\end{align}
where $\boldsymbol{\mu}\in\mathbb{R}^4$ models the mean SDFs, and the covariance matrix $\mathbf{\Sigma}_\phi\in\mathbb{R}^{4\times 4}$ models the topological relation among grid vertices. Both $\boldsymbol{\mu}$ and $\mathbf{\Sigma}_\phi$ are learnable. 
The topological energy over the four grid vertices is given by:
\begin{align}\label{eq:local_top_energy}
    \mathcal{L}_{{\rm top}}(\mathbf{y}_\theta, \boldsymbol{\mu}, \mathbf{\Sigma}_\phi) = \frac{1}{2} (\mathbf{y}_\theta - \boldsymbol{\mu})^T \mathbf{\Sigma}_{\phi}^{-1} (\mathbf{y}_\theta - \boldsymbol{\mu}) + \frac{1}{2}\log|\mathbf{\Sigma}_\phi|,
\end{align}
where the total energy $\mathcal{L}_{\rm top} = \sum_{t\in\mathcal{T}}\mathcal{L}_{{\rm top}, t}$ for the whole tetrahedral grid is a summation over all tetrahedra. \kd{As aforementioned, along with Equation~\ref{eq:local_top_energy}, the $\mathbf{y}_\theta$ and $\boldsymbol{\mu}$ are learnt by using gaussian-to-mesh depth-normal alignment loss functions.}

Key to this formulation is the covariance matrix $\mathbf{\Sigma}_{\phi}$, which can be manually set or learned during optimization. 
\kd{Since the sub-optimality issue of the tetrahedral grid comes from the compromise of local grid vertices within the one-ring neighbor when placing mesh vertices, acknowledging the correlation among grid vertices helps us to identify a pair of vertices that have a strong string force. As a result, the estimation of $\mathbf{\Sigma}_{\phi}$ brings the ability to manipulate the SDF values based on the relation among vertices, enabling accurate surface reconstruction.}
Here, we propose a closed-form solution to learn the SDF via this covariance matrix, which is named Topological SDF, by bringing the construction to the localized SDF at grid vertices \kd{as linear combination of $\mathbf{y}_\theta$ and $\boldsymbol{\mu}$}, which directly expands the degrees of freedom of the SDF.

\paragraph{Closed-form Topological Signed Distance Field Derivation.}  
To obtain the solution form of SDF from Eq.~\eqref{eq:local_top_energy}, we propose an equivalent form of $\mathcal{L}_{{\rm top}}$ using a Reproducing Kernel Hilbert Space~\cite{treves2006topological} (RKHS), which allows us to use the Representer Theorem~\cite{represent_theory} to obtain a closed-form solution. We prove that minimizing the objective function $\mathcal{L}_{\rm top}(\mathbf{y}_\theta)$ is equivalent to the minimization of continuous RKHS topological energy over a tetrahedron. By the Representer Theorem, this formulation yields a unique, closed-form solution continuous field $h^*(x)$ expressed as a linear combination of local tetrahedral vertices' features (Theorem~\ref{thm:osdf}, with proof in Appendix).

\begin{theorem}\label{thm:osdf}[Topological Signed Distance Field]
    Given a tetrahedron $t$ with learnable SDF $\mathbf{y}_\theta\in\mathbb{R}^{4}$, \kd{$\mathbf{\Sigma}_{\phi}$}, and $\boldsymbol{\mu}$, the formal optimization problem on the Reproducing Kernel Hilbert Space (RKHS) $\mathcal{H}_\mathcal{K}$, the minimal continuous energy is exactly equivalent to the discrete Mahalanobis distance over the tetrahedron SDF:
    \begin{align}
        \min_{h} \frac{1}{2} \|h - \boldsymbol{\mu}\|_{\mathcal{H}_K}^2 &= \frac{1}{2} (\mathbf{y}_\theta - \boldsymbol{\mu})^T \mathbf{\Sigma}_{\phi}^{-1} (\mathbf{y}_\theta - \boldsymbol{\mu}) \\
        \text{s.t.} \quad h(\mathbf{x}_i) &= \mathbf{y}_i, \quad \forall i=1,\dots,4, \nonumber
    \end{align}
    where $\boldsymbol{\mu}$ denotes a continuous prior mean. The unique continuous function $h^*(x) = \arg\min_h \frac{1}{2} \|h - m_\theta\|_{\mathcal{H}_K}^2$ solving this variational problem admits the closed-form representation by Representor Theorem~\cite{represent_theory} as follows:
    \begin{align}\label{eq:osdf-solution}
        h^*(x) = m_\theta(x) + \sum_{i=1}^4\Big[\sigma\left(f_\phi(x)^T f_\phi(\mathbf{x}_i)\right)\mathbf{y}_{i}\Big].
    \end{align}
    where $m_\theta(x)$ is denoted as learnable shifting parameter, $f_\phi(x_i)$ is the topological feature at grid vertex $x_i$, and $\sigma$ is the sigmoid function to prevent $f_\phi(x)^T f_\phi(\mathbf{x}_i)$ change the distance sign.
\end{theorem}

By Eq.~\eqref{eq:osdf-solution}, the topologically continuous SDF smoothly interpolates between discrete vertex values while allowing individual edge zero-crossings to adjust dynamically, resolving local geometric frustration in a fully differentiable manner without relying heavily on neural networks. To construct the local covariance $\mathbf{\Sigma}_\phi \in \mathbb{R}^{4 \times 4}$, we define a positive-definite kernel $\mathcal{K}(x, x') = f_\phi(x)^T W f_\phi(x')$, where $W \in \mathbb{R}^{4 \times 4}$ is a symmetric positive-definite metric tensor, and $f_\phi(x): \mathbb{R}^3 \to \mathbb{R}^4$ is a continuous neural coordinate mapping parameterized by $\phi$. The covariance entry between vertices $\mathbf{x}_i$ and $\mathbf{x}_j$ is then $\Sigma_{\phi, ij} = \mathcal{K}(\mathbf{x}_i, \mathbf{x}_j)$.

\paragraph{Adaptation to Gaussians.}
To apply the topological SDF in each training iteration, we first construct the tetrahedral grid over the volume of the scene. Given a set of Gaussians $G_i$ with center $\mu_i$, rotation $R_i$, and scale $S_i$, we assign a learnable normal $n_i$ to each Gaussian and sample two pivots along the Gaussians' normal as $\mu_i + s_in_i$ and  $\mu_i - s_in_i$, where $s_i = \|S_iR_i^Tn_i\|$. From these sampled pivots, we perform Delaunay Triangulation to obtain a set of tetrahedra. At each grid vertex $\mathbf{v}_i$ of the tetrahedral mesh $\mathcal{T}$, we assign learnable SDF $\mathbf{y}_i$ and learnable normal $\mathbf{u}_i$.

During training, the continuous field $h^*(x)$ is evaluated from the current learnable base signed distances $\mathbf{y}_\theta$. The interpolated surface vertices $\mathbf{v}_{t,c}$ and normals $\mathbf{u}_{t,c}$ are extracted along each bipolar edge as follows:
\begin{align}\label{eq:inter_opt_sdf}
    \mathbf{v}_c &= (1 - \hat{\zeta})\mathbf{v}_i + \hat{\zeta}\mathbf{v}_j,\quad\mathbf{u}_c = (1 - \hat{\zeta})\mathbf{u}_i + \hat{\zeta}\mathbf{u}_j \\
    &\text{where}~\hat{\zeta} = \frac{h^*(\mathbf{v}_i)}{h^*(\mathbf{v}_i) - h^*(\mathbf{v}_j)}.
\end{align}
From the interpolated vertices and normals, we construct an explicit triangular surface mesh $\mathcal{M}$, which is used to guide Gaussian optimization via differentiable rasterization.

\paragraph{Properties of Topological SDF.} The Topological SDF formulation has two key properties. First, this continuous topological parameterization dynamically expands the capacity of the local parameter space, enabling candidate zero-crossings along distinct bipolar edges to adjust independently. 

Let ${\rm DoF}_{\rm std}$ and ${\rm DoF}_{\rm top}$ denote the local Degrees of Freedom per tetrahedron for the standard Signed Distance Field and the Topological SDF, respectively. Within a tetrahedron, there are four vertices, meaning that the latent feature $f_\phi(\cdot)\in\mathbb{R}^4$. Under a $4$-dimensional latent feature mapping, the Topological SDF yields ${\rm DoF}_{\rm top} = 20$ (including external learnable parameters), compared to ${\rm DoF}_{\rm std} = 4$ (refer to the proof in Appendix). This expanded capacity provides sufficient candidate vertices adjustments to satisfy conflicting spatial dependencies along distinct bipolar edges, enabling the reconstruction of sharp surface transitions without geometric compromise.

Additionally, the topological SDF formulation improves the training convergence rate. To prove this statement, we leverage a concept called the condition number, a factor that relies on the analysis of the Hessian of the loss function with respect to parameters in Optimization theory~\cite{Nesterov2004}. Basically, a smaller condition number means a faster convergence rate. We provide an external background in the Appendix for a better understanding of these concepts. We formalize our statements in Theorem~\ref{thm:convergence} and provide the corresponding proof in the Appendix.

\begin{theorem}\label{thm:convergence}[Faster Convergence via Hessian Diagonalization]
Let $\mathbf{H}_{\rm std}$ denote the Hessian matrix of the objective function under the standard single-SDF parameterization, and $\mathbf{H}_{\rm top}$ denote the Hessian under the proposed RKHS topological parameterization. If a sharp geometric feature partitions the local vertices into decoupled sets $I$ and $O$ such that the latent mapping forces $\mathcal{K}(\mathbf{x}_i, \mathbf{x}_j) = 0$ for all $\mathbf{x}_i \in I, \mathbf{x}_j \in O$, the condition numbers $\omega$ of the respective Hessians satisfy:
\begin{align}
    \omega(\mathbf{H}_{\rm top}) < \omega(\mathbf{H}_{\rm std}).
\end{align}
Consequently, the topological parameterization achieves a strictly faster linear convergence rate bounded by $\mathcal{O}\left( (1 - 1/\omega(\mathbf{H}_{\rm top}))^\delta \right)$ ($\delta\rightarrow\infty$) compared to the standard representation.
\end{theorem}

\subsection{Energy-based Gaussian Densification Strategy}\label{subsec:mesh-energy}
In the previous section, we derived the closed-form solution of the topological SDF, which allows us to place mesh vertices in a more flexible way to match the target surface, improving mesh-based rendering quality. As aforementioned, we further improve the triangle mesh quality by proposing an energy-based densification via local and global meshing energies. Let us now describe how we regularize the tetrahedral mesh via meshing energies. 

To optimize triangle quality directly during optimization, we minimize the Optimal Delaunay Triangulation (ODT) energy over the tetrahedral grid~\cite{variational_tetrahedral_meshing}. Given a tetrahedral mesh $\mathcal{T}$ defined by vertex pivots $\mathbf{v}_i$, the ODT energy is defined as:
\begin{align}\label{eq:odt_energy}
    \mathcal{L}_{\rm odt}(\mathcal{T}) = \frac{1}{4} \sum_i |\mathbf{v}_i|^2 |\Omega_i| - \int\mathcal{T}|\mathbf{v}|^2d\mathbf{v},
\end{align}
where $|\Omega_i|$ is the volume of the 1-ring neighborhood surrounding vertex $\mathbf{v}_i$. Because the volume integral is constant under a fixed domain boundary $\partial\mathcal{T}$, $\mathcal{L}_{\rm odt}(\mathcal{T})$ acts as an optimal quadratic isotropic regularizer on vertex positions $\mathbf{v}_i$ that minimizes piecewise-linear interpolation error across the domain~\cite{odt}. 

While $\mathcal{L}_{\rm odt}$ provides strong local isotropic regularization, global geometric consistency and uniform dihedral angles across the volume require complementary volumetric regularization. We adopt Weighted Squared Volume Minimization~\cite{wsvm} (WSVM) to penalize volume variance across tetrahedra. 

Approximating the local density $\varrho_t$ at the tetrahedral centroid, the total global volume energy is defined as:
\begin{align}\label{eq:vol_energy}
    \mathcal{L}_{\rm vol}(\mathbf{v}_i) = \mathbb{E}_{\mathbf{v}_i \in \mathcal{T}}\left[\sum_{t=1}^n \varrho_t \overline{V}_t^2(\mathbf{v}_i)\right],
\end{align}
where $\overline{V}_t(\mathbf{v}_i) = \frac{1}{6}(\mathbf{v}_k - \mathbf{v}_l) \cdot ((\mathbf{v}_i - \mathbf{v}_l) \times (\mathbf{v}_j - \mathbf{v}_l))$ and subject to the positive-volume constraint ensuring non-inversion: $(\mathbf{v}_k - \mathbf{v}_l) \cdot ((\mathbf{v}_i - \mathbf{v}_l) \times (\mathbf{v}_j - \mathbf{v}_l)) > 0$.

\begin{figure}
    \centering
    \includegraphics[width=\linewidth]{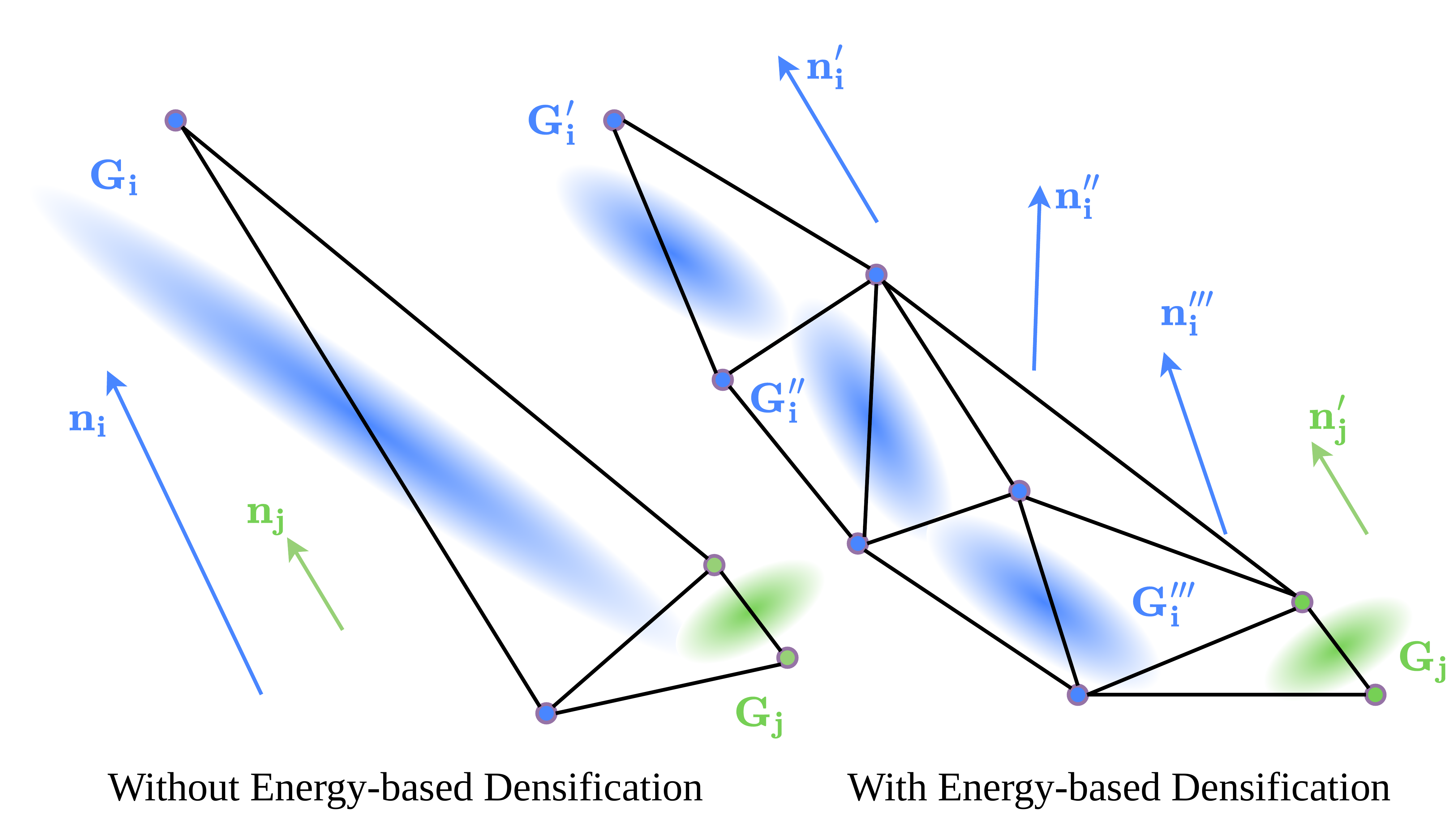}
    \Description{Illustration of energy-based Gaussian densification.}
    \caption{Illustration of the energy-driven Gaussian densification strategy. (Left) Without densification, an elongated Gaussian with large spatial extent induces stretched tetrahedral elements with poor aspect ratios. (Right) With energy-based densification, Gaussians exhibiting high local meshing energy $\mathcal{E}_{\rm odt}$ are dynamically split into smaller, isotropic Gaussians, subdividing the local tetrahedral grid into well-conditioned, uniform elements with aligned normal orientations.}
    \label{fig:densify_strategy}
\end{figure}
Local meshing energy provides a direct metric to identify regions suffering from geometric over-stretching or poor tetrahedral conditioning. By broadcasting local tetrahedral distortion back to individual Gaussians:
\begin{align}\label{eq:energy_densify}
    \mathcal{E}_{\rm odt}(G_i) = \frac{1}{K} \sum_{\mathbf{v} \in \mathcal{P}_{G_i}} \left( \frac{1}{\vert{}\mathcal{T}(\mathbf{v})\vert{}} \sum_{t \in \mathcal{T}(\mathbf{v})} \mathcal{L}_{\rm odt}(\mathcal{T}, t) \right),
\end{align}
where $\mathcal{P}_{G_i}$ is the set of dual pivots sampled from Gaussian $G_i$, and $\mathcal{T}(\mathbf{v})$ denotes the set of tetrahedra incident to vertex $\mathbf{v}$. We define candidate Gaussians for densification as $\mathcal{Q} = \{G_i \mid \mathcal{E}_{\rm odt}(G_i) \ge \tau_\mathcal{E}\}$, where $\tau_\mathcal{E} = \text{Quantile}(\{\mathcal{E}_{\rm odt}(G_i)\}_{i=1}^{N}, \lambda_E)$. Large Gaussians within $\mathcal{Q}$ are split into two smaller child primitives along their major elongation axes. This energy-driven densification adaptively injects degrees of freedom at complex geometric boundaries, transforming elongated elements into more uniform isotropic tetrahedra (Fig.~\ref{fig:densify_strategy}). In practice, we found that densifying based on $\mathcal{L}_{\rm vol}$ does not yield any better result; we therefore perform the proposed densification strategy based only on $\mathcal{L}_{\rm odt}$. 

\begin{algorithm}[t]
\caption{TopSDF End-to-End Training and Optimization Loop}
\label{alg:training_loop}
\begin{algorithmic}[1]
\REQUIRE Multi-view RGB images $\mathcal{I}$, initial Gaussian field $G_0$, warm-up step $\delta_{\rm warm}$, total steps $\delta_{\max}$, grid update interval $\delta_{\rm delaunay}$, densification interval $\delta_{\rm dense}$.
\ENSURE Optimized Gaussian parameters $G^*$, watertight surface mesh $\mathcal{M}^*$.
\STATE $G \leftarrow G_0$
\FOR{$\delta = 1$ \TO $\delta_{\max}$}
  \STATE Differentiably render Gaussian color $\mathbf{C}$, depth $D$, and normal $\tilde{\mathbf{N}}$ from $G$.
  \IF{$\delta > \delta_{\rm warm}$}
    \IF{$\delta = \delta_{\rm warm} + 1$ \OR $(\delta - \delta_{\rm warm}) \pmod{\delta_{\rm delaunay}} = 0$}
      \STATE Sample dual pivots $\mu_i \pm s_i n_i$, where $s_i = \|S_iR_i^Tn_i\|$ (refer to Section~\ref{subsec:ltsdf}).
      \STATE Construct Delaunay grid $\mathcal{T} = \text{Delaunay}(\{\mathbf{v}_i\})$ (refer to Section~\ref{sec:background}).
      \STATE \colorbox{algohighlight}{Initialize SDF field $\mathbf{y}$ and normal field $\mathbf{u}$} at pivots $\mathbf{v}_i$.
    \ENDIF
    \STATE \colorbox{algohighlight}{Compute continuous TopSDF $h^*(\mathbf{x})$} over RKHS (refer to Eq.~\eqref{eq:osdf-solution}).
    \STATE Query bipolar edges ($\mathbf{y}_i \cdot \mathbf{y}_j < 0$) and extract mesh $\mathcal{M}$ with weights $\hat{\zeta}$ (refer to Eq.~\eqref{eq:interpolation} and Eq.~\eqref{eq:inter_opt_sdf}).
    \STATE Compute volume-to-surface alignment losses $\mathcal{L}_{\rm normal}$ and $\mathcal{L}_{\rm depth}$ (refer to Eq.~\eqref{eq:loss_normal_depth}).
    \STATE \colorbox{algohighlight}{Evaluate meshing energies $\mathcal{L}_{\rm odt}(\mathcal{T})$ and $\mathcal{L}_{\rm vol}(\mathcal{T})$} (refer to Eq.~\eqref{eq:odt_energy} and Eq.~\eqref{eq:vol_energy}).
    \IF{$\delta \pmod{\delta_{\rm dense}} = 0$ \AND $\delta < \delta_{\max}$}
      \STATE Compute \colorbox{algohighlight}{energy $\mathcal{E}_{\rm odt}(G_i)$} and split ill-conditioned Gaussians (refer to Eq.~\eqref{eq:energy_densify}).
      \STATE Dynamically update tetrahedral grid $\mathcal{T}$.
    \ENDIF
  \ENDIF
  \STATE Backpropagate composite loss $\mathcal{L}$ (refer to Eq.~\eqref{eq:total_loss}) to update $G$ (and $\mathbf{n}, \mathbf{y}, \mathbf{u}, f_\phi$ when $\delta > \delta_{\rm warm}$).
\ENDFOR
\STATE Extract final watertight surface mesh $\mathcal{M}^*$ from optimized continuous field $h^*$.
\RETURN $G^*, \mathcal{M}^*$
\end{algorithmic}
\end{algorithm}

\subsection{Loss Functions \& Algorithm}

\paragraph{Loss Functions.} We jointly train the Gaussian primitives and their mesh extraction using differentiable rasterization. To align the mesh with the Gaussians, we use differentiable rasterization to render the mesh and compute the depth-normal alignment loss with the Gaussians~\cite{milo}. The depth-normal alignment loss is as follows:
\begin{align}\label{eq:loss_normal_depth}
    \mathcal{L}_{\rm normal} &= \sum_i \left( 1 - \tilde{\mathbf{N}}(\rho)\cdot N_\mathcal{M}(\rho)\right)\\
    \mathcal{L}_{\rm depth} &= \sum_i \log\left(1 + |D(\rho) - D_\mathcal{M}(\rho)|\right),
\end{align}
where $\tilde{\mathbf{N}}(\rho)$, $D(\rho)$ are expected normal, rendered depth from Gaussians, and $N_\mathcal{M}(\rho)$, $D_\mathcal{M}(\rho)$ are mesh rendered normal and depth at pixel $\rho$, respectively. The final loss function is as follows:
\begin{align}\label{eq:total_loss}
    \mathcal{L} = \mathcal{L}_{\rm nvs} + \lambda_{\rm normal}\mathcal{L}_{\rm normal} + \lambda_{\rm depth}\mathcal{L}_{\rm depth} + \lambda_{\rm odt}\mathcal{L}_{\rm odt} + \lambda_{\rm vol}\mathcal{L}_{\rm vol}
\end{align}
\kd{where $\mathcal{L}_{\rm nvs}$ is the novel view synthesis loss, which is the combination of Structural Similarity Index Measure (SSIM) loss~\cite{ssim} and Learned Perceptual Image Patch Similarity loss~\cite{lpips} between the rendered images and ground-truth images}. Otherwise, $\mathcal{L}_{\rm odt}$ is the local meshing energy, and $\mathcal{L}_{\rm vol}$ is the global meshing energy.

\paragraph{Algorithm.} The complete end-to-end training procedure of our framework is formalized in Algorithm~\ref{alg:training_loop}. Throughout all training iterations ($\delta = 1 \dots \delta_{\max}$), Gaussian radiance rendering is continuously optimized via the photometric objective $\mathcal{L}_{\rm nvs}$. During the initial warm-up phase ($\delta \le \delta_{\rm warm}$), Gaussians establish the global scene structure. Once $\delta > \delta_{\rm warm}$, we introduce joint mesh-in-the-loop co-optimization. Specifically, the tetrahedral grid $\mathcal{T}$ is constructed at $\delta = \delta_{\rm warm} + 1$ and periodically reconstructed every $\delta_{\rm delaunay}$ iterations from newly sampled pivots $\mu_i \pm s_i n_i$, where the learnable normal field $\mathbf{u}$ and base SDF $\mathbf{y}$ are initialized. At every subsequent iteration, our continuous Topological SDF $h^*(\mathbf{x})$ is evaluated over the RKHS (refer to Eq.~\eqref{eq:osdf-solution}) to obtain the topological signed distance field, unlocking $20$ local Degrees of Freedom that decouple zero-crossings across shared vertices (refer to Eq.~\eqref{eq:inter_opt_sdf}). We then extract the explicit manifold mesh $\mathcal{M}$, differentiably rasterize mesh depth and normals, and regularize both volume-to-surface alignment (refer to Eq.~\eqref{eq:loss_normal_depth}) and intrinsic meshing energies ($\mathcal{L}_{\rm odt}, \mathcal{L}_{\rm vol}$, refers to Eq.~\eqref{eq:odt_energy} and Eq.~\eqref{eq:vol_energy}). Ill-conditioned Gaussians exhibiting high local distortion are periodically split via energy densification $\mathcal{E}_{\rm odt}(G_i)$ (refer to Eq.~\eqref{eq:energy_densify}). 

Crucially, our framework stands in sharp contrast to existing mesh-in-the-loop pipelines like MILo~\cite{milo}: while MILo relies on discrete scalar parameters ($4$ DoF per tetrahedron) that suffer from severe gradient collisions at shared vertices---forcing reliance on ad-hoc anti-erosion penalties (their $\mathcal{L}_{\rm erosion}$) and periodic ray-cast occupancy checks (their $\mathcal{L}_{\rm interior}$)---our continuous 20-DoF formulation fundamentally eliminates geometric frustration. Furthermore, whereas MILo disregards element shape conditioning, our integrated meshing energies intrinsically guarantee high-quality equilateral elements.

\section{Results}\label{sec:results}

In this section, we present a comprehensive empirical evaluation of our framework across both novel view synthesis and explicit geometric surface reconstruction. We begin by detailing our benchmark datasets and baseline configurations in Section~\ref{subsec:exp_settings}. Next, we evaluate mesh-based novel view rendering on complex unbounded environments in Section~\ref{subsec:mesh_nvs}, and validate surface reconstruction performance in Section~\ref{subsec:surface_recon}. We then demonstrate the adaptability of our pipeline for post-training of a trained set of Gaussians in Section~\ref{subsec:mesh_guidance}, and benchmark triangle quality against traditional remeshing and decimation algorithms in Section~\ref{subsec:remeshing}. Finally, Section~\ref{subsec:ablations} provides systematic ablation studies and empirical validation of our theoretical convergence acceleration.

\subsection{Experimental Settings}\label{subsec:exp_settings}

\paragraph{Training Platform.} All of the experiments in this research are conducted on a single machine with an NVIDIA 5090 GPU ({32GB} VRAM) and {64GB} CPU memory. For the qualitative results of MILo~\cite{milo} only, we utilize a machine with an NVIDIA A6000 GPU ({96GB} VRAM) and {64GB} CPU memory since MILo requires a large amount of memory and resources for training, compared to others.

\paragraph{Datasets.} 
We evaluate our unbounded mesh extraction from Gaussian Splatting~\cite{3dgs} across diverse benchmarks, assessing novel view synthesis and mesh quality on Mip-NeRF 360~\cite{mipnerf360} and Tanks \& Temples~\cite{tnt}. Furthermore, we validate the geometric correctness of our surface reconstructions using both Tanks \& Temples and ScanNet++~\cite{scannetpp}. Finally, supplementary experiments evaluating surface reconstruction on the DTU~\cite{dtu} dataset are detailed in the Appendix.

\paragraph{Comparisons.} We comprehensively compare our method with three categories of 3DGS-based mesh and surface rendering approaches: (i)~\textit{Manifold tetrahedral and volumetric SDF representations}, including Gaussian Opacity Fields~\cite{gof} (GOF), Sorted Opacity Fields~\cite{sof} (SOF), Rasterized Depth GS~\cite{radegs} (RaDe-GS), Geometry Grounded GS~\cite{gggs} (GGGS), Mesh in the Loop Optimization~\cite{milo} (MILo), Gaussian Wrapping~\cite{gw} (GW), and Confidence-Based Mesh Extraction~\cite{come} (CoMe); (ii)~\textit{Primitive and multi-view regularized baselines}, including 2D Gaussian Splatting~\cite{2dgs} (2DGS), Planar-Guided GS Surface Reconstruction~\cite{pgsr} (PGSR), and TopoSurfel~\cite{toposurfel}; and (iii)~\textit{Unconstrained triangle-soup rendering approaches}, including MeshSplatting~\cite{meshsplatting} and Deformable Triangles~\cite{demtrianglesplatting} (DETRIS). To ensure rigorous and fair comparisons, we follow the established evaluation protocols from MILo and GW for mesh novel view synthesis, and CoMe for surface reconstruction.

\begin{table*}[htbp]
\caption{Quantitative evaluation of mesh-based novel view synthesis on MipNeRF-360 and Tanks \& Temples. Methods are categorized into non-manifold triangle soup rendering and manifold complete surface reconstruction approaches. We report rendering fidelity metrics (PSNR$\uparrow$, SSIM$\uparrow$, LPIPS$\downarrow$), extracted mesh vertex counts (\#Verts$\downarrow$), and total training duration in minutes (Time$\downarrow$). The best scores among manifold reconstruction methods are formatted in \textbf{bold}, and second-best are \underline{underlined}.\label{tab:mesh_nvs}}
\centering
\setlength{\tabcolsep}{5.5pt}
\begin{tabular}{lccccccccccc}
\toprule
\multirow{2}{*}{\textbf{Method}} & \multirow{2}{*}{\textbf{Venues}} & \multicolumn{5}{c}{\textbf{MipNeRF360}} & \multicolumn{5}{c}{\textbf{Tank \& Temples}} \\
                                 &                                  & PSNR$\uparrow$ & SSIM$\uparrow$  & LPIPS$\downarrow$ & \#Verts$\downarrow$ & Time$\downarrow$ & PSNR$\uparrow$ & SSIM$\uparrow$  & LPIPS$\downarrow$ & \#Verts$\downarrow$ & Time$\downarrow$ \\ \midrule
\multicolumn{12}{c}{\textit{Non-manifold Triangle Soup - Mesh-based Rendering}} \\ \midrule                                   
MeshSplatting & CVPR2026 & 24.78 & 0.310 & 0.7280 & 3M & 48 & 20.52 & 0.287 & 0.745 & 2M  & 48 \\
DETRIS & ECCV2026 & 27.70 & 0.8280 & 0.1830 & 12M & 55 & 23.31 & 0.8600 & 0.1330 & 8M & 55 \\ \midrule
\multicolumn{12}{c}{\textit{Manifold Complete Surface Mesh - Surface Reconstruction}} \\ \midrule
2DGS & SIGGRAPH2024 & 15.36 & 0.4987 & 0.4749 & \textbf{4.31M} & 12 & 14.23 & 0.5697 & 0.4854 & 16.39M & 12\\
PGSR & TVCG2024 & 15.63 & 0.5834 & 0.4509 & 22.78M & 45 & 14.55 & 0.5956 & 0.4819 & 11.79M & 45 \\
MILo                             & TOG2025                          & 25.07          & 0.7339          & 0.3096            & 6.73M               & 150              & 21.20          & 0.6908          & 0.3782            & {\ul 4.36M}         & 160              \\
TopoSurfel & TOG2026 & 17.28 & 0.5961 & 0.4402 & 5.96M & 98 & 16.77 & 0.6326 & 0.4702 & 5.02M & 98 \\
Ours (small)                     &                                  & {\ul 25.10}    & 0.7475          & 0.2970            & {\ul 5.86M}         & {\ul 25}         & 21.20          & 0.7016          & 0.3450            & \textbf{4.16M}      & {\ul 25}         \\ \midrule
GOF                              & TOG2024                          & 24.25          & 0.7017          & 0.3454            & 32.80M              & 37               & 20.10          & 0.6475          & 0.4073            & 11.63M              & 40               \\
RaDe-GS                          & TOG2026                          & 24.84          & 0.7291          & 0.3128            & 30.85M              & \textbf{20}      & 20.70          & 0.6767          & 0.3876            & 10.06M              & \textbf{20}      \\
GGGS                             & SIGASIA2026                             & 24.55          & 0.7386          & 0.2986            & 43.46M              & \textbf{20}      & 20.70          & 0.6912          & 0.3734            & 21.86M              & \textbf{20}      \\
GW                               & ECCV2026                             & {\ul 25.10}    & {\ul 0.7520}    & \textbf{0.2890}   & 11.79M              & 30               & 20.98          & 0.7004          & 0.3674            & 5.80M               & 30               \\
CoMe                             & ECCV2026                         & 23.64          & 0.6610          & 0.3320            & 18.43M              & 40               & 21.19          & 0.6830          & \textbf{0.3450}   & 13.06M              & 40               \\ \midrule
Ours                             &                                  & \textbf{25.96} & \textbf{0.7655} & \textbf{0.2890}   & 11.82M              & 50               & \textbf{21.72} & \textbf{0.7125} & {\ul 0.3510}      & 6.32M               & 50               \\ \bottomrule
\end{tabular}
\end{table*}



\begin{figure*}
    \centering
    \includegraphics[width=\linewidth]{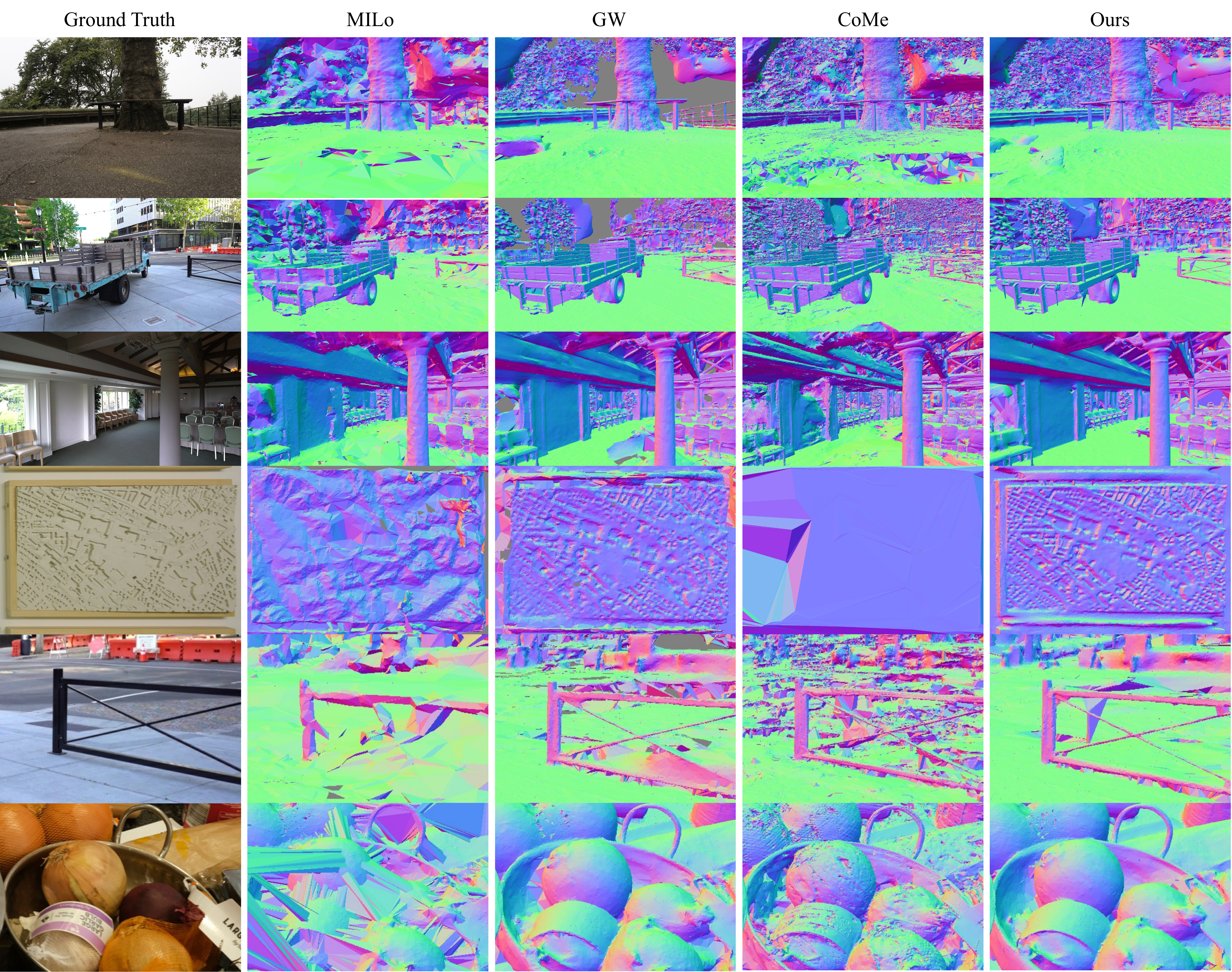}
    \Description{Qualitative surface geometry comparisons across diverse scenes.}
    \caption{Qualitative comparison of extracted surface geometry across diverse indoor and outdoor scenes. The leftmost column displays the ground truth RGB reference image. Subsequent columns display the rendered surface normal maps extracted by MILo~\cite{milo}, GW~\cite{gw}, CoMe~\cite{come}, and our framework, respectively. Additional qualitative scene comparisons are provided in the Appendix. Zoom in for details.}
    \label{fig:main}
\end{figure*}

\begin{figure*}
    \centering
    \includegraphics[width=\linewidth]{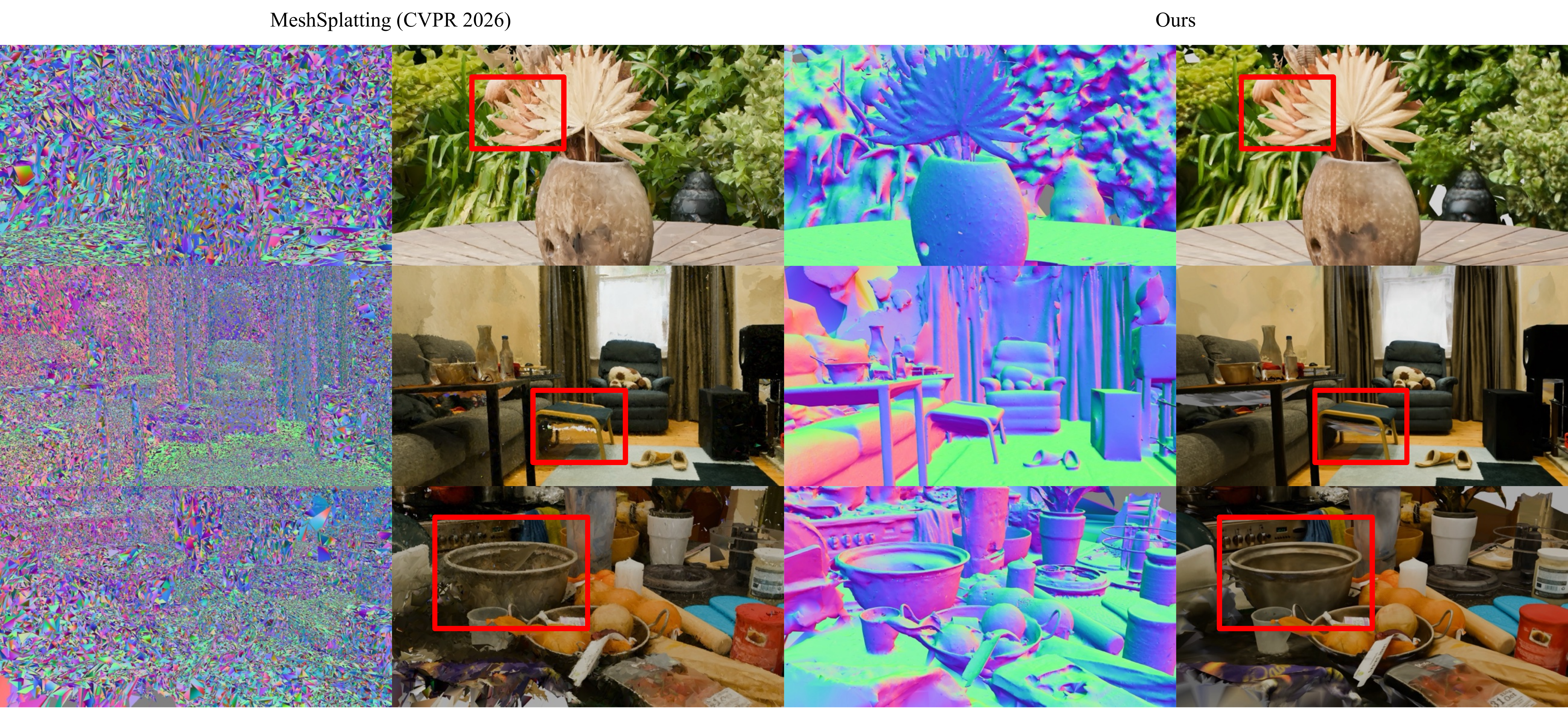}
    \Description{Visual comparison between MeshSplatting and our manifold surface meshes.}
    \caption{Visual comparison between non-manifold triangle soup rendering (MeshSplatting~\cite{meshsplatting}, left) and our continuous manifold surface mesh extraction (Ours, right). For each method, we show the rendered surface normal map alongside the corresponding novel view RGB image across three challenging scenes. Zoom in for details.}
    \label{fig:vsmsplat}
\end{figure*}


\subsection{Mesh-based Novel View Synthesis}\label{subsec:mesh_nvs}

We follow MILo~\cite{milo} and GW~\cite{gw} to perform quantitative comparisons among methods on mesh novel view synthesis (reported in Tab.~\ref{tab:mesh_nvs}). We evaluate on the standard predefined test camera views from both unbounded indoor/outdoor scenes in MipNeRF-360~\cite{mipnerf360} and large-scale environments in Tanks \& Temples~\cite{tnt}. To render the colored mesh, we leverage Nvdiffrast~\cite{Laine2020diffrast} for differentiable rasterization and compute PSNR, SSIM, and LPIPS against the ground-truth test images. 
To conduct a fair comparison, we keep the default settings of all baselines.

As shown in Tab.~\ref{tab:mesh_nvs}, our method achieves substantial improvements in rendering quality across all benchmarks. On MipNeRF-360, our full model achieves \textbf{25.96 dB} PSNR and \textbf{0.7655} SSIM, outperforming the previous state-of-the-art GW by \textbf{+0.86 dB} in PSNR and +0.0135 in SSIM, while maintaining the best perceptual fidelity (LPIPS of 0.2890). On Tanks \& Temples, our approach similarly establishes a new state-of-the-art with \textbf{21.72 dB} PSNR (+0.74 dB over GW and +0.52 dB over MILo) and \textbf{0.7125} SSIM.

An insightful pattern emerges when analyzing the performance trajectory of existing manifold reconstruction methods. Prior tetrahedral grid approaches—including GOF, RaDe-GS, GGGS, MILo, GW, and CoMe—consistently encounter an empirical performance ceiling around 24.5--25.1 dB on MipNeRF-360. Notably, simply scaling vertex density does not resolve this bottleneck: for instance, GGGS allocates 43.46M vertices, and GOF uses 32.80M vertices, yet both remain capped at 24.55 dB and 24.25 dB, respectively. This empirical observation strongly corroborates our theoretical analysis in Section~\ref{subsec:ltsdf}: the standard single-scalar SDF parameterization (4 DoF per tetrahedron) imposes severe geometric frustration when multiple bipolar edges share conflicting gradient updates at boundaries. By elevating the representation to our 20-DoF Topological SDF, our method successfully decouples cross-boundary constraints, breaking this fundamental representation bottleneck without requiring an excessive number of vertices.

Regarding computational efficiency and geometry scale, while our full model requires 50 minutes of training for 11.82M vertices on MipNeRF-360, it delivers the highest fidelity. To address resource-constrained scenarios, we also introduce a lightweight variant, \textit{Ours (small)}, which matches the rendering performance of GW (25.10 dB PSNR) with only 5.86M vertices in just 25 minutes of training. Compared to MILo (which requires 150 minutes to optimize 6.73M vertices), \textit{Ours (small)} achieves a $\mathbf{6\times}$ training speedup compared MILO~\cite{milo}, and superior SSIM (0.7475 vs. 0.7339) with a smaller vertex footprint.

We perform qualitative evaluations against manifold tetrahedral baselines (MILo, GW, and CoMe) in Fig.~\ref{fig:main} and unconstrained triangle-soup representations (MeshSplatting~\cite{meshsplatting}) in Fig.~\ref{fig:vsmsplat}. Several distinct geometric patterns emerge across these comparisons:

\paragraph{Scene-Scale vs. Object-Centric Fidelity.} In large-scale unbounded scenes (e.g., park environments, truck scenes, and meeting rooms in Fig.~\ref{fig:main}), prior methods struggle with depth transitions across spatial scales: GW leaves large missing surface voids behind major foreground occluders (such as tree trunks and truck cabs), MILo fragments background floor boundaries, and CoMe introduces heavy high-frequency roughness across planar grounds. In contrast, our framework maintains leak-free surface geometry, reconstructing unbroken ground planes and clean architectural elements (columns, ceiling beams, and walls). Concurrently, in object-centric spaces (e.g., fruit bowls, relief maps, and indoor furniture), while competing methods either over-smooth curvature (MILo) or produce noisy, crusty surfaces (CoMe), our approach recovers smooth, continuous curvature on organic shapes while preserving crisp boundary silhouettes.

\paragraph{Preservation of Thin Structures and Micro-Geometries.} Thin geometric structures and high-frequency reliefs represent critical failure modes for conventional discrete SDF representations due to geometric frustration at shared vertices. As observed in Fig.~\ref{fig:main}, MILo over-smooths thin structures, snapping thin X-braced fence bars into disconnected fragments and washing out intricate city map reliefs into generic undulations; GW severely dilates and distorts thin fence posts; and CoMe accumulates chaotic clouds of floating triangle shards or suffers from degenerate planar collapse on relief maps. In contrast, our Topological SDF successfully decouples local zero-crossings, extracting intact, sharp, and continuous thin bars (unbroken fence crossbars and thin metallic bowl rims) and resolving high-frequency micro-reliefs (sharp street grids and rectangular frame borders) without geometric dilation.

\paragraph{Manifold Surface Continuity vs. Triangle-Soup Glitter.} In Fig.~\ref{fig:vsmsplat}, we contrast our extracted surface meshes against MeshSplatting~\cite{meshsplatting}. While MeshSplatting produces plausible novel view RGB renderings from standard viewpoints, its surface normal maps reveal an unoriented, chaotic "triangle glitter"—a disjointed soup of millions of floating, unaligned facets with zero topological continuity. Such non-manifold representations are fundamentally unsuitable for downstream geometric modeling, physics simulation, or collision detection. In stark contrast, our framework recovers globally smooth, topologically coherent surface normals across complex environments (e.g., smooth leaves and table planks in garden scenes, structured armchair cushions and walls in living rooms, and clean curvature on kitchen bowls and bottles), while matching or exceeding photorealistic RGB rendering. 

\subsection{Surface Reconstruction}\label{subsec:surface_recon}

\begin{table}[htbp]
\caption{Quantitative evaluation of surface reconstruction accuracy on Tanks \& Temples and ScanNet++. We report F1-scores ($\uparrow$) evaluated against laser-scanned ground truth point clouds across outdoor and indoor environments. The highest scores are formatted in \textbf{bold}, and second-highest are \underline{underlined}.\label{tab:surrecon}}
\centering
\setlength{\tabcolsep}{1.8pt}
\begin{tabular}{@{}lcccccccc@{}}
\toprule
            & GOF        & SOF  & RaDe-GS & MILo       & GGGS       & GW            & CoMe          & \textbf{Ours} \\ \midrule
\multicolumn{9}{c}{Tank \& Temples}                                                                                \\ \midrule
Barn        & 0.40       & 0.45 & 0.42    & 0.40       & 0.57       & \textbf{0.64} & 0.53          & {\ul 0.66}    \\
Caterpillar & 0.23       & 0.20 & 0.22    & 0.24       & 0.41       & 0.41          & {\ul 0.47}    & \textbf{0.50} \\
Courthouse  & 0.10       & 0.07 & 0.12    & 0.11       & 0.10       & 0.19          & \textbf{0.33} & {\ul 0.30}    \\
Ignatius    & 0.55       & 0.68 & 0.66    & 0.72       & {\ul 0.79} & 0.77          & \textbf{0.78} & \textbf{0.78} \\
Meetingroom & 0.19       & 0.15 & 0.22    & 0.14       & 0.36       & {\ul 0.37}    & {\ul 0.37}    & \textbf{0.40} \\
Truck       & 0.41       & 0.39 & 0.45    & 0.43       & 0.50       & 0.50          & {\ul 0.63}    & \textbf{0.64} \\ \midrule
Mean        & 0.31       & 0.32 & 0.35    & 0.34       & 0.45       & 0.48          & {\ul 0.52}    & \textbf{0.55} \\ \midrule
\multicolumn{9}{c}{ScanNet++}                                                                                       \\ \midrule
5a269ba6fe  & {\ul 0.62} & 0.60 & -       & {\ul 0.62} & -          & \textbf{0.67} & \textbf{0.67} & \textbf{0.67} \\
08bbbdcc3d  & 0.67       & 0.68 & -       & 0.67       & -          & {\ul 0.73}    & 0.72          & \textbf{0.74} \\
39f36da05b  & 0.61       & 0.61 & -       & 0.62       & -          & {\ul 0.65}    & {\ul 0.65}    & \textbf{0.66} \\
c263dfbf0   & 0.67       & 0.67 & -       & 0.62       & -          & 0.70          & {\ul 0.71}    & \textbf{0.73} \\
ef18cf0708  & 0.50       & 0.49 & -       & 0.54       & -          & {\ul 0.56}    & 0.55          & \textbf{0.57} \\
fb564c935d  & 0.66       & 0.62 & -       & 0.66       & -          & {\ul 0.69}    & 0.68          & \textbf{0.72} \\ \midrule
Mean        & 0.62       & 0.61 & -       & 0.62       & -          & {\ul 0.66}    & {\ul 0.66}    & \textbf{0.67} \\ \bottomrule
\end{tabular}
\end{table}

\begin{table}[htbp]
\caption{Quantitative evaluation of foreground surface reconstruction on Tanks \& Temples. We report per-scene and mean F1-scores ($\uparrow$) across six object-centric benchmark scenes: Barn, Caterpillar (Ca), Courthouse (Co), Ignatius (Ig), Meetingroom (MR), and Truck (Tr). Best results are formatted in \textbf{bold} and second-best are \underline{underlined}.\label{tab:surrecon-foreground}}
\centering
\begin{tabular}{@{}lccccccc@{}}
\toprule
     & Barn          & Ca   & Co    & Ig      & MR   & Tr         & Avg           \\ \midrule
2DGS & {\ul 0.48}    & 0.24          & 0.13          & 0.51          & 0.18          & 0.46          & 0.33          \\
PGSR & \textbf{0.65} & {\ul 0.45}    & \textbf{0.24} & \textbf{0.81} & {\ul 0.33} & 0.63    & \textbf{0.52} \\
TopoSurfel & \textbf{0.65} & 0.43    & {\ul 0.22} & {\ul 0.77} & \textbf{0.34} & {\ul 0.64} & \textbf{0.52} \\
Ours & \textbf{0.65} & \textbf{0.52} & 0.17    & 0.75    & 0.28    & \textbf{0.65} & {\ul 0.50}    \\ \bottomrule
\end{tabular}
\end{table}

To further evaluate the geometric fidelity of our extracted meshes, we follow standard protocols~\cite{gw, come} across unbounded scenes from Tanks \& Temples~\cite{tnt} and complex indoor spaces from ScanNet++~\cite{scannetpp}. We jointly optimize the 3D Gaussians with the signed distance fields on each scene and extract the definitive explicit surface mesh. Accuracy is evaluated by computing the F1-score against ground-truth laser-scanned point clouds.

\paragraph{Unbounded Surface Reconstruction.} As reported in Tab.~\ref{tab:surrecon}, our framework achieves the highest overall surface reconstruction fidelity across both benchmarks, setting a new state-of-the-art mean F1-score of \textbf{0.55} on Tanks \& Temples (+0.03 over CoMe, +0.07 over GW, and +0.21 over MILo) and \textbf{0.67} on ScanNet++. 

On Tanks \& Temples, our approach demonstrates consistent performance gains across diverse environments. Specifically, in the \textit{Caterpillar} scene (0.50 F1) and \textit{Truck} scene (0.64 F1), our method outperforms prior tetrahedral baselines by substantial margins (+0.09 and +0.14 over GW and GGGS, respectively). In \textit{Barn} (0.66 F1) and \textit{Meetingroom} (0.40 F1), our method also achieves the top reconstruction scores, while matching CoMe on \textit{Ignatius} (0.78 F1). On the \textit{Courthouse} scene, our approach reaches 0.30 F1, outperforming earlier methods like GW (0.19) and MILo (0.11). Similarly, on ScanNet++, which features complex indoor scenes with intricate furniture structures and planar surfaces, our method achieves the highest or tied F1-score across all evaluated scenes, showing robust reconstruction capability in dense indoor spaces.

\begin{figure*}[t]
    \centering
    \includegraphics[width=\linewidth]{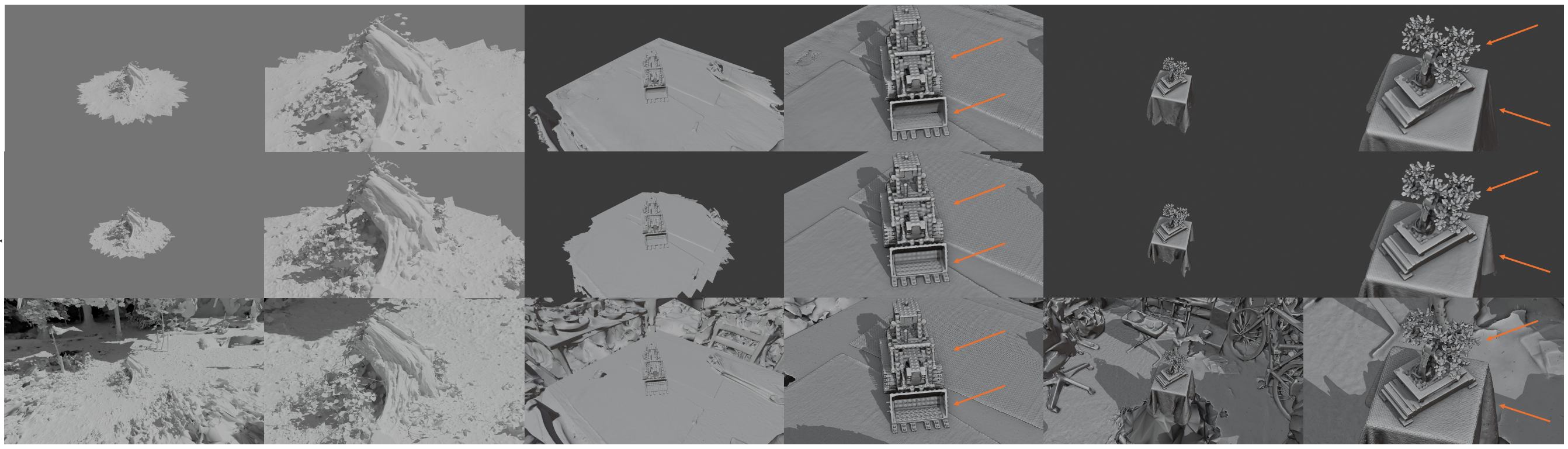}
    \Description{Bound vs Unbound Mesh Extraction.}
    \caption{Bounded vs Unbounded Mesh Extraction. Qualitative comparisons between the mesh extraction sheme between our approach and 2DGS and TopoSurfel. While we can obtain full scene mesh, 2DGS and TopoSurfel focus only on foreground mesh extraction, which leads to reduction in mesh novel-view systhesis quality. Zoom in for details.}
    \label{fig:bvsub}
\end{figure*}

\paragraph{Foreground Surface Reconstruction.} We benchmark foreground-specific mesh extraction on Tanks \& Temples against 2DGS~\cite{2dgs}, PGSR~\cite{pgsr}, and TopoSurfel~\cite{toposurfel} in Tab.~\ref{tab:surrecon-foreground}. 2DGS achieves an average F1-score of 0.33, struggling with noisy geometry from unconstrained disk primitives. PGSR and TopoSurfel achieve an average F1-score of 0.52 by incorporating dense multi-view stereo geometric constraints. Our method achieves a competitive average F1-score of \textbf{0.50} without relying on specialized foreground heuristics, and outperforms PGSR and TopoSurfel on challenging scenes such as \textit{Caterpillar} (\textbf{0.52} vs. 0.45 \& 0.43) and \textit{Truck} (\textbf{0.65} vs. 0.63 \& 0.64), while matching PGSR and TopoSurfel on \textit{Barn} (\textbf{0.65}). Even though on average PGSR and TopoSurfel have a slightly better reconstruction performance compared to ours, those approaches cannot reconstruct unbounded surfaces (refer to Fig.~\ref{fig:bvsub}). 

\subsection{External Mesh Guidance}\label{subsec:mesh_guidance}

\begin{table*}[htbp]
\caption{Evaluation of plug-and-play mesh guidance fine-tuning on pretrained Gaussian representations (GW~\cite{gw} and CoMe~\cite{come}). We report rendering fidelity metrics (PSNR$\uparrow$, SSIM$\uparrow$, LPIPS$\downarrow$), extracted mesh vertex counts (\#Vert), and triangle element quality metrics ($Q_{\rm avg}\uparrow$, AR$\uparrow$, RR$\uparrow$, TR$\uparrow$, ReR$\downarrow$, AD$\downarrow$, $V_{567}\uparrow$) before and after fine-tuning with our pipeline. Improvements are highlighted in \textbf{bold}.\label{tab:sur-recon-mesh-guide}}
\centering
\setlength{\tabcolsep}{4.5pt}
\renewcommand{\arraystretch}{0.8}
\begin{tabular}{lccccccccccc}
\toprule
            & PSNR$\uparrow$ & SSIM$\uparrow$  & LPIPS$\uparrow$ & \#Vert & $Q_{\rm avg}$$\uparrow$ & AR$\uparrow$  & RR$\uparrow$  & TR$\uparrow$  & ReR$\downarrow$     & AD$\downarrow$ & $V_{\rm 567}\uparrow$ (\%) \\ \midrule
GW          & 25.10          & 0.7520          & 0.2890          & 11.79  & 0.46                    & 0.37          & 0.24          & 0.45          & 4.31                & 0.58           & 58.9                        \\
GW + Ours   & \textbf{25.52} & \textbf{0.7598} & \textbf{0.2802} & 11.84  & \textbf{0.58}           & \textbf{0.46} & \textbf{0.32} & \textbf{0.51} & {\ul \textbf{4.22}} & \textbf{0.56}  & \textbf{70.8}               \\ \midrule
CoMe        & 23.64          & 0.6610          & 0.3320          & 24.32  & 0.45                    & 0.35          & 0.25          & 0.43          & 4.76                & 0.62           & 60.2                        \\
CoMe + Ours & \textbf{24.86} & \textbf{0.7372} & 0.3320          & 24.61  & \textbf{0.56}           & \textbf{0.45} & \textbf{0.32} & \textbf{0.48} & 4.76                & 0.62           & \textbf{71.1}               \\ \bottomrule
\end{tabular}
\end{table*}
To evaluate the modular adaptability of TopSDF, we investigate whether our continuous topological parameterization and variational meshing energies can serve as an effective geometric refinement module for existing pre-trained 3DGS representations. Specifically, we initialize our framework directly with the optimized Gaussians, sampled pivots, and base SDF fields from Gaussian Wrapping (GW)~\cite{gw} and CoMe~\cite{come}, fine-tuning solely for a brief $1{,}000$ iterations under our composite objective (Eq.~\eqref{eq:total_loss}).

As shown in Tab.~\ref{tab:sur-recon-mesh-guide}, fine-tuning pre-trained representations with our pipeline decisively boosts both rendering fidelity and geometric quality across all benchmarks. When plugged into GW, our framework elevates PSNR from $25.10\text{ dB}$ to $25.75\text{ dB}$ ($+0.65\text{ dB}$), while simultaneously improving average triangle quality ($Q_{\rm avg}: 0.46 \to 0.58$). Similarly, applied to CoMe, our method achieves a $+1.58\text{ dB}$ surge in PSNR ($23.64 \to 25.22\text{ dB}$) and elevates triangle quality ($Q_{\rm avg}: 0.45 \to 0.57$). Although energy-driven densification introduces a modest increase in vertex footprint to resolve localized geometric tension, these results confirm that our topological formulation and variational energies generalize seamlessly across diverse Gaussian initializations, reliably unlocking higher photorealism and superior geometric conditioning in minutes.

\subsection{Comparisons with Remeshing and Decimation}\label{subsec:remeshing}

\begin{table*}[htbp]
\caption{Comparison of mesh novel view synthesis and triangle element quality against post-processing remeshing pipelines. We compare our directly optimized meshes against raw baselines (GW~\cite{gw}, CoMe~\cite{come}) and post-processed variants using Isotropic Remeshing (RM) and Quadric Decimation (DM). Metrics include rendering quality (PSNR$\uparrow$, SSIM$\uparrow$, LPIPS$\downarrow$), mesh scale (\#Vert$\downarrow$), additional post-processing runtime (Extra Time$\downarrow$), and geometric quality indicators ($Q_{\rm avg}\uparrow$, AR$\uparrow$, RR$\uparrow$, TR$\uparrow$, ReR$\downarrow$, AD$\downarrow$, $V_{567}\uparrow$). Best results are formatted in \textbf{bold} and second-best are \underline{underlined}.\label{tab:mesh-quality}}
\centering
\setlength{\tabcolsep}{4.5pt}
\begin{tabular}{@{}lcccccccccccc@{}}
\toprule
             & PSNR           & SSIM            & LPIPS           & \#Vert          & Extra Time & $Q_{\rm avg}\uparrow$ & AR$\uparrow$  & RR$\uparrow$  & TR$\uparrow$  & ReR$\downarrow$ & AD$\downarrow$ & $V_{\rm 567}\uparrow$ (\%) \\ \midrule
GW           & 25.10          & 0.7520          & 0.2890          & \textbf{11.79M} & -          & 0.46                  & 0.37          & 0.24          & 0.45          & 4.31            & 0.58           & 58.9                        \\
CoMe         & 23.64          & 0.6610          & 0.3320          & 18.43M          & -          & 0.45                  & 0.35          & 0.25          & 0.43          & 4.76            & 0.62           & 60.2                        \\
GW + RM      & {\ul 25.10}    & {\ul 0.7520}    & {\ul 0.2890}    & 24.32M          & +20 mins   & {\ul 0.60}            & \textbf{0.51} & \textbf{0.33} & {\ul 0.53}    & \textbf{4.17}   & {\ul 0.44}     & 85.9                        \\
GW + RM + DM & 24.86          & 0.7268          & 0.2810          & 15.32M          & +20 mins   & 0.52                  & 0.36          & 0.29          & 0.47          & 4.28            & 0.50           & \textbf{72.2}               \\ \midrule
Ours         & \textbf{25.96} & \textbf{0.7655} & \textbf{0.2890} & {\ul 11.82M}    & -          & \textbf{0.62}         & {\ul 0.49}    & \textbf{0.33} & \textbf{0.55} & {\ul 4.21}      & \textbf{0.42}  & {\ul 71.6}                  \\ \bottomrule
\end{tabular}
\end{table*}

\begin{table*}[htbp]
\caption{Ablation study over the four core components of our framework: continuous Topological SDF parameterization ($h^\prime\rightarrow h^*$), local Delaunay meshing energy ($\mathcal{L}_{\rm odt}$), global volume meshing energy ($\mathcal{L}_{\rm vol}$), and $\mathcal{L}_{\rm odt}$-based densification strategy ($\mathcal{E}_{\rm odt}(G)$). Each row represents a specific active combination ($\checkmark$). We report rendering metrics (PSNR$\uparrow$, SSIM$\uparrow$, LPIPS$\downarrow$), mesh scale (\#Vert$\downarrow$), and geometric mesh quality metrics ($Q_{\rm avg}\uparrow$, AR$\uparrow$, RR$\uparrow$, TR$\uparrow$, ReR$\downarrow$, AD$\downarrow$, $V_{567}\uparrow$). Cells indicating degraded performance relative to baseline are highlighted in blue ({\color{blue!15}{$\blacksquare$}}).\label{tab:abs-top-energy}}
\centering
\setlength{\tabcolsep}{5pt}
\begin{tabular}{@{}ccccccccccccccc@{}}
\toprule
\multicolumn{1}{l}{$h^\prime\rightarrow h^*$} & \multicolumn{1}{l}{$\mathcal{L}_{\rm odt}$} & \multicolumn{1}{l}{$\mathcal{L}_{\rm vol}$} & \multicolumn{1}{l}{$\mathcal{E}_{\rm odt}(G)$} & PSNR$\uparrow$ & SSIM$\uparrow$ & LPIPS$\downarrow$ & \#Vert$\downarrow$ & $Q_{\rm avg}$$\uparrow$ & AR$\uparrow$ & RR$\uparrow$ & TR$\uparrow$ & ReR$\downarrow$ & AD$\downarrow$ & $V_{\rm 567}\uparrow$ (\%) \\ \midrule
                                             &                                             &                                             &                                                  & 24.88          & 0.7339          & 0.3096            & \textbf{11.72M}    & 0.50                    & 0.37         & 0.15         & 0.45         & 4.31            & 0.58           & 60.9                        \\ \midrule
$\checkmark$                                 &                                             &                                             &                                                  & 25.58          & 0.7562          & 0.2978            & \textbf{11.72M}    & \cellcolor{blue!15}0.47 & \cellcolor{blue!15}0.35 & 0.17         & \cellcolor{blue!15}0.40 & \cellcolor{blue!15}4.51 & \cellcolor{blue!15}0.62 & 65.8                        \\
                                             & $\checkmark$                                 &                                             &                                                  & \cellcolor{blue!15}24.42 & \cellcolor{blue!15}0.7295 & \cellcolor{blue!15}0.3125 & \textbf{11.72M}    & 0.55                    & 0.42         & 0.20         & 0.48         & 4.25            & 0.48           & 63.2                        \\
                                             &                                             & $\checkmark$                                 &                                                  & 25.32          & 0.7482          & 0.3055            & \cellcolor{blue!15}{\ul 11.80M} & 0.58            & 0.45         & 0.23         & 0.51         & 4.20            & 0.45           & 65.7                        \\ \midrule
$\checkmark$                                 & $\checkmark$                                 &                                             &                                                  & 25.16          & 0.7511          & 0.2985            & \textbf{11.72M}    & 0.57                    & 0.44         & 0.20         & 0.47         & 4.22            & 0.55           & 66.0                        \\
                                             & $\checkmark$                                 & $\checkmark$                                 &                                                  & 24.92          & 0.7454          & 0.3022            & \textbf{11.72M}    & 0.59                    & 0.47         & 0.24         & 0.48         & 4.20            & 0.46           & 67.2                        \\
                                             & $\checkmark$                                 &                                             & $\checkmark$                                     & 25.66          & 0.7482          & 0.2967            & \cellcolor{blue!15}11.82M & \underline{0.61}        & \underline{0.48}& 0.26         & 0.52         & 4.18            & \textbf{0.38}  & 67.8                        \\ \midrule
$\checkmark$                                 & $\checkmark$                                 & $\checkmark$                                 &                                                  & 25.66          & {\ul 0.7578}    & 0.2943            & \textbf{11.72M}    & 0.59                    & 0.45         & 0.25         & 0.51         & {\ul 4.17}      & {\ul 0.40}     & 67.5                        \\
$\checkmark$                                 & $\checkmark$                                 &                                             & $\checkmark$                                     & {\ul 25.77}    & {\ul 0.7578}    & {\ul 0.2896}      & \cellcolor{blue!15}{\ul 11.80M} & \underline{0.61}        & 0.47         & {\ul 0.30}   & {\ul 0.53}   & \textbf{4.16}   & {\ul 0.40}     & {\ul 69.2}                  \\
                                             & $\checkmark$                                 & $\checkmark$                                 & $\checkmark$                                     & 25.46          & 0.7496          & 0.2933            & \cellcolor{blue!15}11.82M & \textbf{0.62}           & \textbf{0.49}& \textbf{0.33}& \textbf{0.55}& \textbf{4.16}   & {\ul 0.40}     & \textbf{71.6}               \\ \midrule
$\checkmark$                                 & $\checkmark$                                 & $\checkmark$                                 & $\checkmark$                                     & \textbf{25.96} & \textbf{0.7655} & \textbf{0.2890}   & \cellcolor{blue!15}11.82M & \textbf{0.62}           & \textbf{0.49}& \textbf{0.33}& \textbf{0.55}& 4.21            & 0.42           & \textbf{71.6}               \\ \bottomrule
\end{tabular}
\end{table*}
\begin{figure}
    \centering
    \includegraphics[width=\linewidth]{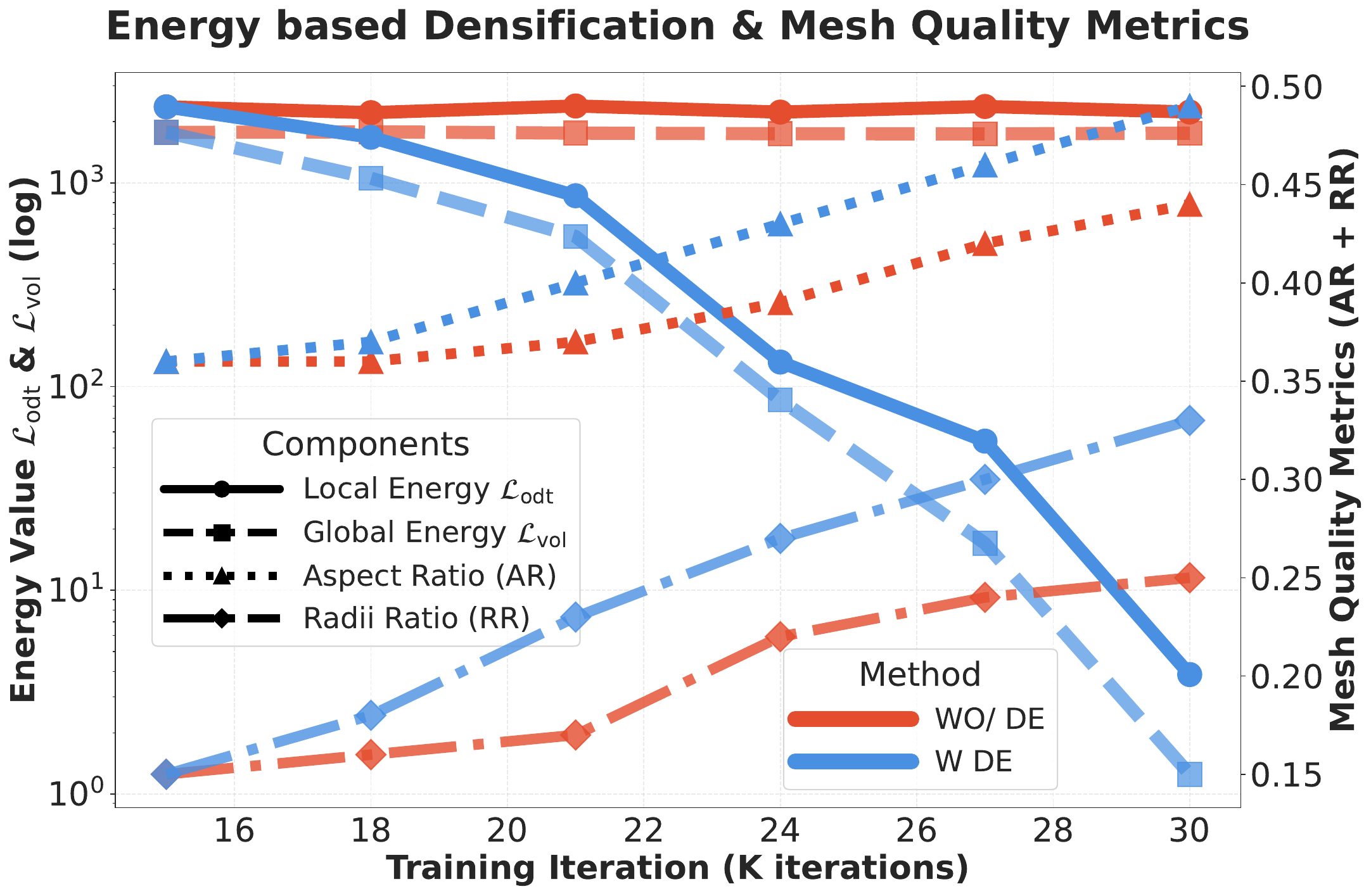}
    \Description{Optimization trajectories of meshing energies and mesh quality metrics.}
    \caption{Optimization trajectories of local ($\mathcal{L}_{\rm odt}$) and global ($\mathcal{L}_{\rm vol}$) meshing energies alongside mesh quality metrics across training iterations. Left vertical axis (logarithmic scale) tracks energy values for models trained without (red lines) and with (blue lines) energy-based densification (DE). Right vertical axis plots the geometric mesh conditioning metrics: Aspect Ratio (AR, dotted lines) and Radii Ratio (RR, dash-dotted lines).}
    \label{fig:densify}
\end{figure}

To quantitatively evaluate mesh quality, we adopt standard remeshing metrics~\cite{remeshing}: triangle quality~\cite{quality} $\mathcal{Q}(w) = \frac{6}{\sqrt{3}} \frac{A_w}{S_w E_w}$, aspect ratio (AR)~\cite{ar} $\frac{e_w}{E_w}$, radii ratio (RR)~\cite{radiiratio} $\frac{r_w}{R_w}$, triangle regularity (TR)~\cite{triangleregularity} $\frac{2\sqrt{3}r_w}{E_w}$, radius edge ratio (ReR)~\cite{rer} $\frac{R_w}{e_w}$, angle deviation~\cite{angledeviation} ${\rm AD}(\mathcal{M}) = \mathbb{E}_{\gamma\in\mathcal{M}}[|\gamma - \gamma_{\rm opt}|]$, and vertex regularity $V_{567}$~\cite{isotropicremeshing, vertexregularity} (percentage of valence 5, 6, or 7 vertices). Here $A_w, S_w, e_w, E_w, R_w, r_w, \gamma$, and $\gamma_{\rm opt}=\frac{\pi}{3}$ denote face area, semi-perimeter, shortest edge length, longest edge length, circumradius, inradius, interior angle, and optimal equilateral angle, respectively.

In Tab.~\ref{tab:mesh-quality}, we evaluate the triangle element quality and rendering fidelity of our extracted meshes against raw baseline meshes from GW~\cite{gw} and CoMe~\cite{come}, as well as post-processed variants using Isotropic Remeshing (RM)~\cite{isotropicremeshing, pymeshlab} and Quadric Edge Collapse Decimation (DM).

As shown in Tab.~\ref{tab:mesh-quality}, raw surface meshes extracted via standard Marching Tetrahedra (GW and CoMe) exhibit severe geometric degradation in triangle shape quality. Because their optimization focuses solely on photometric error, the resulting meshes contain numerous needle-like slivers and degenerate triangles, reflected by an average quality of $Q_{\rm avg} \leq 0.46$, and a high angle deviation ($\text{AD} \geq 0.58$).

Applying post-processing remeshing (GW + RM) improves element regularity ($Q_{\rm avg} = 0.60$, $\text{AD} = 0.44$, and $V_{567} = 85.9\%$). However, this improvement incurs substantial practical costs: it requires an additional 20 minutes of post-processing computation and more than doubles the mesh size to \textbf{24.32M} vertices ($+106\%$ increase over raw GW), while offering zero improvement in rendering fidelity (25.10 dB PSNR). Furthermore, applying subsequent decimation (GW + RM + DM) to reduce the vertex count back to 15.32M inevitably disrupts geometric alignment, degrading rendering fidelity to \textbf{24.86 dB} PSNR (a $-0.24\text{ dB}$ drop) and causing triangle quality to deteriorate ($Q_{\rm avg}$ drops back to $0.52$, and $\text{AD}$ increases to $0.50$).

In contrast, our framework resolves this fundamental trade-off by directly integrating local ($\mathcal{L}_{\rm odt}$) and global ($\mathcal{L}_{\rm vol}$) meshing energies into the joint optimization. Our method produces the highest average triangle quality ($Q_{\rm avg} = \mathbf{0.62}$), the highest triangle regularity ($\text{TR} = \mathbf{0.55}$), and the lowest angle deviation ($\text{AD} = \mathbf{0.42}$, indicating triangles closest to ideal $60^\circ$ equilateral shapes). Crucially, our approach achieves this superior mesh conditioning while simultaneously delivering state-of-the-art novel view rendering (\textbf{25.96 dB} PSNR and \textbf{0.7655} SSIM) with \textbf{zero post-processing runtime overhead} and a compact vertex footprint (11.82M vertices).

\subsection{Ablation Studies \& Empirical Analysis}\label{subsec:ablations}

\paragraph{Topological SDF \& Meshing Energy.} To evaluate the individual and compound contributions of each component in our framework, we systematically ablate four key modules in Tab.~\ref{tab:abs-top-energy}: the continuous Topological SDF parameterization ($h^\prime\rightarrow h^*$), the local Delaunay meshing energy ($\mathcal{L}_{\rm odt}$), the global volume meshing energy ($\mathcal{L}_{\rm vol}$), and the $\mathcal{L}_{\rm odt}$-based densification strategy ($\text{DE}$). 

Activating the continuous Topological SDF ($h^*$) in isolation substantially elevates novel view rendering fidelity over the baseline, increasing PSNR from $24.88\text{ dB}$ to $25.58\text{ dB}$ and SSIM from $0.7339$ to $0.7562$. However, because $h^*$ optimizes unconstrained zero-crossings against photometric loss, triangle shapes stretch to minimize pixel residuals, degrading mesh quality across almost all metrics ($Q_{\rm avg}$ drops from $0.50$ to $0.47$, $\text{AR}$ to $0.35$, $\text{TR}$ to $0.40$, $\text{ReR}$ rises to $4.51$, and angle deviation $\text{AD}$ worsens to $0.62$). Conversely, enforcing local Delaunay meshing energy $\mathcal{L}_{\rm odt}$ alone suppresses degenerate sliver elements ($\text{AD}$ drops to $0.48$, and $\text{AR}$ increases to $0.42$), but over-constrains the fixed tetrahedral grid without densification, causing rendering performance to degrade (PSNR drops to $24.42\text{ dB}$ and SSIM to $0.7295$). In comparison, global volume energy $\mathcal{L}_{\rm vol}$ alone offers moderate simultaneous gains in both rendering ($25.32\text{ dB}$ PSNR) and geometry ($Q_{\rm avg} = 0.58, \text{AR} = 0.45$). 

Combining local Delaunay energy with energy-driven densification ($\mathcal{L}_{\rm odt} + \text{DE}$) yields a decisive performance surge, elevating PSNR by $+1.24\text{ dB}$ over $\mathcal{L}_{\rm odt}$ alone (reaching $25.66\text{ dB}$), improving $Q_{\rm avg}$ to $0.61$, and achieving the lowest isolated angle deviation ($\text{AD} = \mathbf{0.38}$). Introducing $h^*$ into this pairing ($h^* + \mathcal{L}_{\rm odt} + \text{DE}$) further accelerates rendering fidelity ($25.77\text{ dB}$ PSNR, $0.7578$ SSIM, and $0.2896$ LPIPS) while preserving high element regularities ($Q_{\rm avg} = 0.61, \text{RR} = 0.30, V_{567} = 69.2\%$), whereas coupling both meshing energies with densification ($\mathcal{L}_{\rm odt} + \mathcal{L}_{\rm vol} + \text{DE}$) maximizes geometric conditioning ($Q_{\rm avg} = \mathbf{0.62}, \text{AR} = \mathbf{0.49}, \text{RR} = \mathbf{0.33}, \text{TR} = \mathbf{0.55}, V_{567} = \mathbf{71.6\%}$). Ultimately, our full framework combining all four components ($h^* + \mathcal{L}_{\rm odt} + \mathcal{L}_{\rm vol} + \text{DE}$) achieves the global optimum across both domain objectives, delivering state-of-the-art novel view rendering (\textbf{25.96 dB} PSNR, \textbf{0.7655} SSIM, and \textbf{0.2890} LPIPS) while strictly matching more regular triangle condition ($Q_{\rm avg} = \mathbf{0.62}$, $\text{AR} = \mathbf{0.49}$, $\text{RR} = \mathbf{0.33}$, $\text{TR} = \mathbf{0.55}$, and $V_{567} = \mathbf{71.6\%}$).

\paragraph{Meshing Energy-based Densification.} We further investigate the convergence dynamics of our local ($\mathcal{L}_{\rm odt}$) and global ($\mathcal{L}_{\rm vol}$) meshing energies in Fig.~\ref{fig:densify}, comparing training with and without the energy-based densification strategy ($\text{DE}$). Without densification, the meshing energies plateau at a high loss magnitude ($>10^3$) throughout training, reflecting optimization stagnation where the fixed tetrahedral resolution cannot resolve localized geometric distortions. In contrast, applying energy-based densification dynamically allocates spatial degrees of freedom to poorly conditioned tetrahedra, causing both local and global energy losses to drop steeply by over three orders of magnitude (down to $\sim 4 \times 10^0$ and $\sim 10^0$, respectively). As depicted on the right axis of Fig.~\ref{fig:densify}, this rapid energy minimization directly drives substantial, sustained improvements in geometric quality, accelerating Aspect Ratio ($\text{AR}$) from $0.35$ to $\mathbf{0.49}$ and more than doubling Radii Ratio ($\text{RR}$) from $0.15$ to $\mathbf{0.33}$. 

\paragraph{Convergence Rate Analysis.} According to Theorem~\ref{thm:convergence}, under the circumstances that the Hessian matrix $\mathbf{H}$ is symmetric and positive definite, the convergence rate of gradient descent is bounded by the condition number $\omega(\mathbf{H})$. We show that empirically, this setting not only makes the implementation less complicated but also accelerates the convergence rate. 

To validate this theoretical result, we isolate all components that are not related to SDF learning and record the mesh novel view synthesis PSNR across optimization iterations for both Single SDF and Topological SDF. Note that this implementation is different from MILo~\cite{milo} in that we do not use the anti-erosion loss $\mathcal{L}_{\rm erosion}$ and interior loss $\mathcal{L}_{\rm interior}$, which are designed to mitigate the gradient collision problem. 

\begin{figure}[t]
    \centering
    \includegraphics[width=\linewidth]{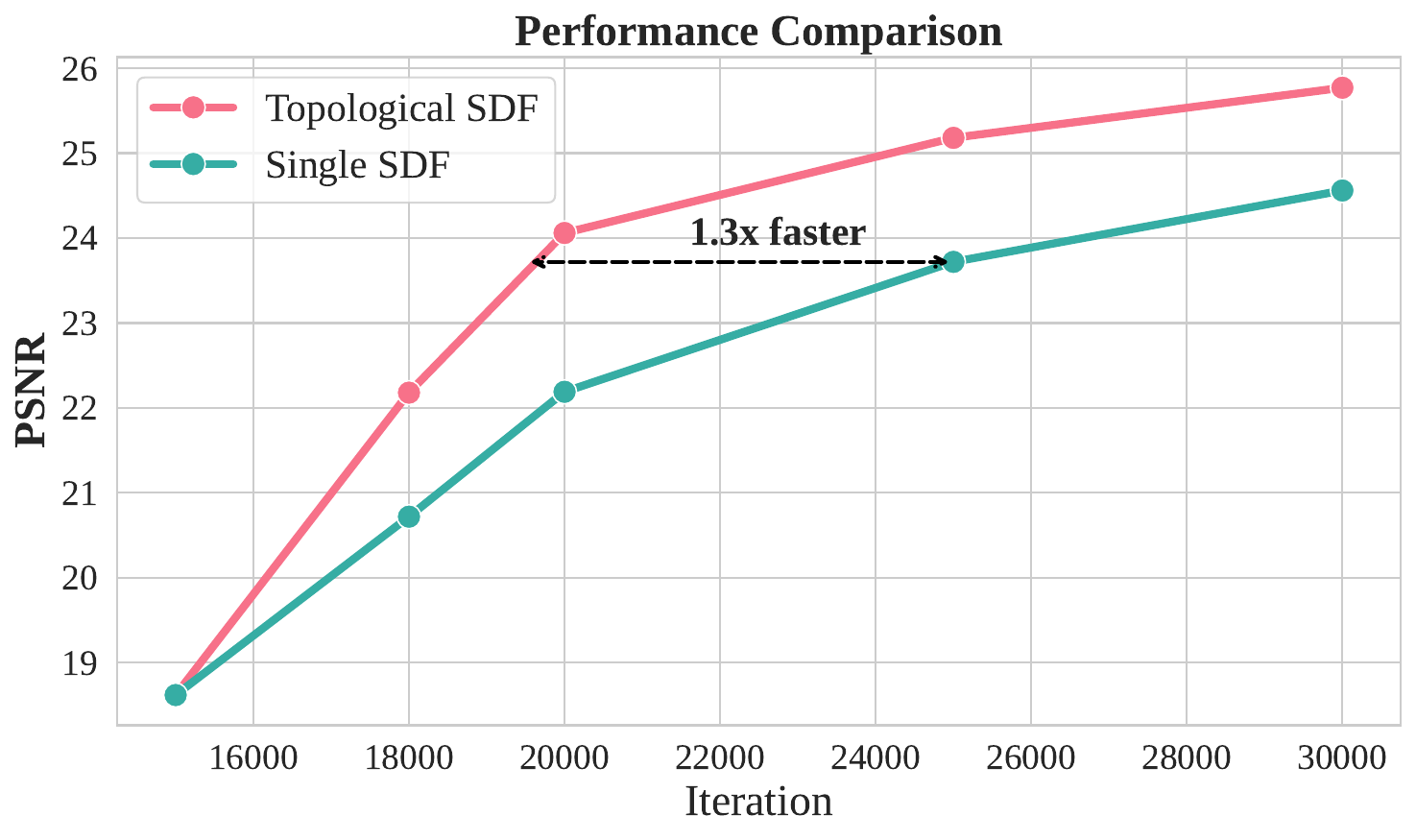}
    \Description{Convergence rate comparison between Topological SDF and Single SDF.}
    \caption{Convergence Rate Analysis. Performance comparison of mesh novel view synthesis (PSNR) over optimization iterations on MipNeRF-360. We plot the convergence trajectories of Topological SDF (pink) against the standard Single SDF (teal), starting after the initial 15,000-step Gaussian warm-up phase. The dashed arrow indicates the convergence acceleration required to reach equivalent rendering fidelity.}
    \label{fig:convergence_rate}
\end{figure}

As shown in Fig.~\ref{fig:convergence_rate}, following the initial 15,000-iteration Gaussian warm-up phase, both representations begin joint mesh optimization from an identical baseline of $18.63\text{ dB}$. Topological SDF immediately demonstrates a significantly steeper optimization trajectory, reaching $22.18\text{ dB}$ at iteration 18,000 and $24.08\text{ dB}$ at iteration 20,000. In contrast, Single SDF requires 25,000 iterations to attain a comparable rendering level ($23.72\text{ dB}$), demonstrating a direct $\mathbf{1.3\times}$ speedup in optimization convergence. Furthermore, while Single SDF plateaus around $24.56\text{ dB}$ at iteration 30,000 due to gradient stagnation across shared vertices, Topological SDF continues to ascend, reaching a superior final performance of $\mathbf{25.77\text{ dB}}$ ($+1.21\text{ dB}$ improvement). This empirical progression directly validates Theorem~\ref{thm:convergence}: by block-diagonalizing the optimization Hessian through orthogonalized latent feature covariance, Topological SDF fundamentally eliminates cross-boundary gradient interference, thereby lowering the condition number $\omega(\mathbf{H})$ and enabling faster, higher-fidelity convergence without requiring heuristic anti-erosion penalties.

\section{Limitations}
While our framework significantly advances explicit surface reconstruction, it still possesses certain limitations. First, calculating the topological SDF via the latent feature covariance introduces additional computational overhead during training compared to standard scalar SDF representations. Evaluating the RKHS continuous meshing energy requires querying the kernel function across multiple vertices, which increases the memory footprint slightly and limits the maximum batch size of tetrahedra that can be processed simultaneously. Finally, although the 20-DoF parameterization provides massive flexibility to resolve local geometric frustration, the reconstruction is still fundamentally bound by the global resolution of the underlying tetrahedral grid. Extremely thin structures or high-frequency details that fall below the spatial discretization limit of the grid may still be under-represented, even with our meshing energy-based densification strategy. Future work could explore more memory-efficient kernel formulations or adaptive hierarchical tetrahedral grids to further push the boundaries of high-fidelity explicit surface extraction.

\section{Conclusion}
In this paper, we revisit the problem of extracting a watertight triangle mesh from 3D Gaussian Splatting. 
We derive a closed-form solution for SDF estimates on tetrahedron vertices via learning topological correlation, and leverage variational meshing energies to regularise the joint optimization of 3D Gaussians and triangle mesh extraction, supported by an energy-based Gaussian densification strategy. 
Our experiments on multiple datasets show that we not only improve the mesh-based rendering quality but also the triangle mesh quality.

In future work, we aim to include geometry priors to guide the geometry of Gaussians to stay semantically aligned to corresponding regions in the scene. Yet, a promising avenue is to extend the proposed framework to address mesh construction from dynamic Gaussians. Last but not least, current approaches take a large amount of time for training, compared to naive 3DGS, which we aim to improve in future research.

\clearpage
\bibliographystyle{ACM-Reference-Format}
\bibliography{acmart}

\clearpage
\appendix



\section{Extended Backgrounds}\label{sec:extended_backgrounds}

\paragraph{Reproducing Kernel Hilbert Spaces.} An RKHS $\mathcal{H}_\mathcal{K}$ over a spatial domain $\mathcal{X} \subseteq \mathbb{R}^3$ is a Hilbert space of real-valued functions equipped with an inner product $\langle \cdot, \cdot \rangle_{\mathcal{H}_\mathcal{K}}$ and a unique symmetric, positive-definite reproducing kernel $\mathcal{K}: \mathcal{X} \times \mathcal{X} \to \mathbb{R}$~\cite{represent_theory}. This space is characterized by two properties: 1) Reproducing Property: For any $x \in \Omega$, the kernel function $\mathcal{K}(\cdot, x)$ belongs to $\mathcal{H}_\mathcal{K}$, and any continuous function $h \in \mathcal{H}_\mathcal{K}$ satisfies the reproducing identity: $h(x) = \langle h, \mathcal{K}(\cdot, x) \rangle_{\mathcal{H}_\mathcal{K}}$. 2) Smoothness: For any function $h\in\mathcal{H}_\mathcal{K}$, the evaluation at two points $x, x^\prime\in \mathcal{X}$ satisfies the inequality $|h(x) - h(x^\prime)|\leq\|f\|_{\mathcal{H}_\mathcal{K}}\cdot d_\mathcal{K}(x, x^\prime)$, where $d_\mathcal{K}(x, x^\prime) = \sqrt{\mathcal{K}(x, x^\prime)+\mathcal{K}(x^\prime, x^\prime) - 2\mathcal{K}(x, x^\prime)}$, which is a bounded constant that depends on $x$ and the choice of kernel.

\paragraph{Representer Theorem.} In this paper, we leverage the Representer Theorem to derive the closed-form of the topological SDFs. In statistical learning theory, the Representer Theorem~\cite{represent_theory} states that a minimizer $g^*$ of a regularized empirical risk function defined over a Reproducing Kernel Hilbert Space (RKHS) can be treated as a finite linear combination of input points' kernel products. Formally, consider a positive-definite real-valued kernel $\mathcal{K}: \mathcal{X}\times\mathcal{X}\rightarrow\mathbb{R}$ on the non-empty set $\mathcal{X}$ over RKHS $\mathcal{H}_K$ and given a training set $(x_1, y_1), \dots, (x_n, y_n)\in\mathcal{X}\times\mathbb{R}$, a strictly increasing real-valued function $\ell:[0, \infty]\rightarrow\mathbb{R}$, and an arbitrary error function $\mathcal{L}:(\mathcal{X}\times\mathbb{R}^2)^n$, the minimizer is as follows:
\begin{align}
    g^* = \underset{g\in\mathcal{H}_{\mathcal{K}}}{{\rm argmin}} \{\mathcal{L}((x_1, y_1, g(x_1), \dots, x_n, y_n, g(x_n)))\} + \ell(\|g\|),
\end{align}
admits the following representation form:
\begin{align}
    g^*(\cdot) = \sum_{i=1}^n \beta_i\mathcal{K}(\cdot, x_i).
\end{align}

\paragraph{Gershgorin Circle Theorem.} The Gershgorin Circle theorem~\cite{gershgorin1931uber} bounds the spectrum (eigenvalues) of a square matrix based on its entries. Let $\mathbf{A} \in \mathbb{C}^{n \times n}$ be a square matrix with entries $a_{ij}$. For each row $i$, the Gershgorin disc is defined as a closed region in the complex plane centered at the diagonal element $a_{ii}$ with a radius $r_i = \sum_{j \neq i} |a_{ij}|$ representing the sum of the absolute values of the off-diagonal entries. The theorem dictates that every eigenvalue of $\mathbf{A}$ must lie within at least one of these discs. 

A critical corollary arises when the matrix $\mathbf{A}$ is a real symmetric matrix. Because the eigenvalues of a real symmetric matrix are guaranteed to be strictly real, the complex Gershgorin discs collapse into closed intervals on the real number line: $[a_{ii} - r_i, a_{ii} + r_i]$. Consequently, the absolute minimum and maximum eigenvalues of a real symmetric matrix can be strictly bounded by observing the maximum span of these intervals. Specifically, the maximum eigenvalue is bounded by $\lambda_{\max} \leq \max_i (a_{ii} + r_i)$, and the minimum eigenvalue is bounded by $\lambda_{\min} \geq \min_i (a_{ii} - r_i)$. The inequality can be represented as follows:
\begin{align}
    |\lambda - a_{ii}| \leq \sum_{i\neq j}|a_{ij}|
\end{align}

\paragraph{Convergence Analysis of Gradient Descent.} For an optimization solved via first-order methods (e.g., Gradient Descent), the convergence rate towards a stationary point $\mathbf{y}^*$ is bounded by the condition number $\omega$ of the Hessian matrix $\mathbf{H} = \nabla^2 \mathcal{L}$:
\begin{align}
    \omega(\mathbf{H}) = \frac{\lambda_{\max}(\mathbf{H})}{\lambda_{\min}(\mathbf{H})}, 
\end{align}
where $\lambda_{\max}(\mathbf{H}$) and $\lambda_{\min}(\mathbf{H}$) are the largest and the smallest eigenvalues of $\mathbf{H}$. The distance to the stationary point at step $\delta$ decays geometrically at a rate bounded by:
\begin{align}
    || \mathbf{y}_\theta^{(\delta)} - \mathbf{y}_\theta^* || \leq \mathcal{O}\left( \left( 1 - \frac{1}{\omega(\mathbf{H})} \right)^\delta \right),
\end{align}
where $\mathbf{y}_\theta^{(\delta)}$ is the SDF state at step $\delta$. A larger $\omega$ implies slow, stiff convergence, while a smaller $\omega$ yields fast linear convergence. This convergence rate is crucial, as it will help us to analyze the bound of the convergence rate of our method compared to others.

\section{Mathematical Proofs for Theorem and Theoretical Analysis}\label{sec:proofs}

\subsection{Proof for Topological Sign Distance Field Theorem}

\begin{proof}
    First, we prove that given a kernel $\mathcal{H}_\mathcal{K}$, evaluating the continuous energy of our local field $h^*(\mathbf{x}_i)$ collapses exactly into the discrete Mahalanobis energy of tetrahedra $t$. Formally, we prove that
    \begin{align}
        \frac{1}{2} \|h^*(x) - \boldsymbol{\mu}\|_{\mathcal{H}_\mathcal{K}}^2 = \frac{1}{2} (\mathbf{y}_\theta - \boldsymbol{\mu})^T \mathbf{\Sigma}_{\phi}^{-1} (\mathbf{y}_\theta - \boldsymbol{\mu}).
    \end{align}
    We begin with the continuous energy of the minimizer residual $g^*(x) = h^*(x) - \boldsymbol{\mu}$. Substituting $g^*(x)$ into the continuous energy of the local signed distance field, the left-hand side becomes:
    \begin{align}
        \frac{1}{2} \|g^*\|_{\mathcal{H}_\mathcal{K}}^2 = \frac{1}{2} \langle g^*, g^* \rangle_{\mathcal{H}_\mathcal{K}}.
    \end{align}
    According to the Representer theorem~\cite{represent_theory}, the residual function $g^*(x)$ can be represented as $g^*(\cdot) = \sum_{i=1}^4 \beta_i \mathcal{K}(\cdot, \mathbf{x}_i)$; therefore,
    \begin{align}
        \frac{1}{2} \|g^*\|_{\mathcal{H}_\mathcal{K}}^2 = \frac{1}{2} \left\langle \sum_{i=1}^4 \beta_i \mathcal{K}(\cdot, \mathbf{x}_i), \sum_{j=1}^4 \beta_j \mathcal{K}(\cdot, \mathbf{x}_j) \right\rangle_{\mathcal{H}_\mathcal{K}},
    \end{align}
    where $\beta_i\in\mathbb{R}$ for all $1 \leq i \leq 4$. Expanding the summations via the bilinearity of the inner product:
    \begin{align}
        \frac{1}{2} \|g^*\|_{\mathcal{H}_\mathcal{K}}^2 = \frac{1}{2} \sum_{i=1}^4 \sum_{j=1}^4 \beta_i \beta_j \langle \mathcal{K}(\cdot, \mathbf{x}_i), \mathcal{K}(\cdot, \mathbf{x}_j) \rangle_{\mathcal{H}_\mathcal{K}}.
    \end{align}
    By the reproducing property of the RKHS ($\langle \mathcal{K}(\cdot, \mathbf{x}_i), \mathcal{K}(\cdot, \mathbf{x}_j) \rangle_{\mathcal{H}_\mathcal{K}} = \mathcal{K}(\mathbf{x}_i, \mathbf{x}_j) = (\mathbf{\Sigma}_\phi)_{ij}$), we have:
    \begin{align}
        \frac{1}{2} \|g^*\|_{\mathcal{H}_\mathcal{K}}^2 = \frac{1}{2} \sum_{i=1}^4 \sum_{j=1}^4 \beta_i \beta_j (\mathbf{\Sigma}_\phi)_{ij} = \frac{1}{2} \boldsymbol{\beta}^T \mathbf{\Sigma}_\phi \boldsymbol{\beta}.
    \end{align}
    Since $g^*(\cdot) = \sum_{i=1}^4 \beta_i \mathcal{K}(\cdot, \mathbf{x}_i)$, the residual at vertex $j$ is $h(\mathbf{x}_j) - \boldsymbol{\mu}_j = \sum_{i=1}^4 \beta_i \mathcal{K}(\mathbf{x}_j, \mathbf{x}_i)$ for all $j \in \{1, \dots, 4\}$. Writing this linear system in matrix form gives $\mathbf{y}_\theta - \boldsymbol{\mu} = \mathbf{\Sigma}_\phi \boldsymbol{\beta}$. Because $\mathcal{K}$ is strictly positive-definite, $\mathbf{\Sigma}_\phi$ is invertible, yielding $\boldsymbol{\beta} = \mathbf{\Sigma}_\phi^{-1} (\mathbf{y}_\theta - \boldsymbol{\mu})$. Substituting $\boldsymbol{\beta}$ back into the energy:
    \begin{align}
        \frac{1}{2} \|g^*\|_{\mathcal{H}_\mathcal{K}}^2 &= \frac{1}{2} \left( \mathbf{\Sigma}_\phi^{-1} (\mathbf{y}_\theta - \boldsymbol{\mu}) \right)^T \mathbf{\Sigma}_\phi \left( \mathbf{\Sigma}_{\phi}^{-1} (\mathbf{y}_\theta - \boldsymbol{\mu}) \right) \notag \\
        &= \frac{1}{2} (\mathbf{y}_\theta - \boldsymbol{\mu})^T \mathbf{\Sigma}_{\phi}^{-1} \mathbf{\Sigma}_\phi \mathbf{\Sigma}_{\phi}^{-1} (\mathbf{y}_\theta - \boldsymbol{\mu}) \\
        &= \frac{1}{2} (\mathbf{y}_\theta - \boldsymbol{\mu})^T \mathbf{\Sigma}_{\phi}^{-1} (\mathbf{y}_\theta - \boldsymbol{\mu}).
    \end{align}
    Let $k(x) = [\mathcal{K}(x, \mathbf{x}_i), \mathcal{K}(x, \mathbf{x}_j), \mathcal{K}(x, \mathbf{x}_k), \mathcal{K}(x, \mathbf{x}_l)]^T$. The minimizer of topoplogical signed distance field $h^*(x)$ is then derived as:
    \begin{align}
        g^*(x) &= k(x)^T \mathbf{\Sigma}_{\phi}^{-1} (\mathbf{y}_\theta - \boldsymbol{\mu}) \\
        \Leftrightarrow h^*(x) - \boldsymbol{\mu} &= k(x)^T \mathbf{\Sigma}_{\phi}^{-1} (\mathbf{y}_\theta - \boldsymbol{\mu}) \\
        \Leftrightarrow h^*(x) &= \boldsymbol{\mu} + k(x)^T \mathbf{\Sigma}_{\phi}^{-1} (\mathbf{y}_\theta - \boldsymbol{\mu}). \notag
    \end{align}
    To maintain the kernel $\mathcal{K}$ as a positive-definite matrix, we set $\mathbf{\Sigma}_\phi$ to be a diagonal matrix, yielding:
    \begin{align}
        h^*(x) &= \boldsymbol{\mu} + k(x)^T{\rm diag}(\{\mathcal{K}(x, x_i)\}_{i=1}^4)(\mathbf{y}_\theta - \boldsymbol{\mu}) \notag \\ 
        &= \boldsymbol{\mu} + k(x)^T(k(x) \odot (\mathbf{y}_\theta - 
        \boldsymbol{\mu})) \\
        &= \boldsymbol{\mu} + \sum_{i=1}^4\mathcal{K}(x,\mathbf{x}_i)(\mathbf{y}_i - \boldsymbol{\mu}_{t, i}) \\
        &= \boldsymbol{\mu} + \sum_{i=1}^4 f_\phi(x)^Tf_\phi(x_i)(\mathbf{y}_i - \boldsymbol{\mu}_{t, i}).
    \end{align}
    Under $W=\mathbb{I}$, the kernel evaluation collapses to the feature dot product. To strictly maintain the sign of the distance, we employ the sigmoid function $\sigma$ on the scale component $f_\phi(x)^Tf_\phi(x_i)$. In practice, we absorb all the shifting component into a learnable shifting parameter $m_\theta(x)$. The simplified topological signed distance field is then:
    \begin{align}
        h^*(x) = m_\theta(x) + \sum_{i=1}^4 \sigma\left(f_\phi(x)^T f_\phi(\mathbf{x}_i)\right)\mathbf{y}_i.
    \end{align}
\end{proof}

\subsection{Proof for Topological Degrees of Freedom for Accurate Surface Reconstruction}\label{sec:appendix_top_flex}
\begin{proof}
    Recall that in Marching Tetrahedra, the tetrahedron $t$ is defined by four vertices $\mathbf{v}_i$ for $i \in \{1, 2, 3, 4\}$. To obtain the triangle mesh $\mathcal{M}$, we need to query all bipolar edges and perform interpolation to obtain the triangle vertices. The interpolation is defined as follows:
    \begin{align}
        \mathbf{v}_c = (1 - \hat{\zeta})\mathbf{v}_i + \hat{\zeta}\mathbf{v}_j, \quad \text{where}~\hat{\zeta} = \frac{h(\mathbf{v}_i)}{h(\mathbf{v}_i) - h(\mathbf{v}_j)},
    \end{align}
    where $h$ is the signed distance field. The interpolation is only valid when $h(\mathbf{v}_i) \cdot h(\mathbf{v}_j) < 0$. 
    
    \textbf{Topological SDF (${\rm DoF}_{\rm top} = 20$):} 
    In our framework, rather than assigning a single isolated scalar to each vertex, we formulate the vertex values as a continuous solution $h^*(x)$ parameterized by the local topological state. Specifically, the signed distance value at a discrete vertex is defined by an abstract function $h^*(\mathbf{v}_i) = \psi(\mathbf{y}_\theta, F_\phi, \mathbf{v}_i)$, where $F_\phi = [f_\phi(\mathbf{v}_1), \dots, f_\phi(\mathbf{v}_4)]^T \in \mathbb{R}^{4 \times 4}$ represents the $4$-dimensional latent feature vectors evaluated at the 4 vertices of the tetrahedron. This formulation depends not only on the local scalar states $\mathbf{y}_\theta \in \mathbb{R}^4$ but fundamentally on the topological features in $F_\phi$. 
    
    Consequently, the interpolation weight $\hat{\zeta}_{ij}$ along any bipolar edge is no longer dictated by two isolated scalars. Instead, it is governed by the full topology of the tetrahedra:
    \begin{align}
        \hat{\zeta}_{ij} = \frac{\psi(\mathbf{y}_\theta, F_\phi, \mathbf{v}_i)}{\psi(\mathbf{y}_\theta, F_\phi, \mathbf{v}_i) - \psi(\mathbf{y}_\theta, F_\phi, \mathbf{v}_j)}.
    \end{align}
    Because the feature network outputs a $4$-dimensional vector for each of the 4 vertices, it introduces $16$ independent local topological parameters. Combined with the 4 scalar parameters in $\mathbf{y}_\theta$, the placement of any candidate vertex in the Topological SDF is dynamically guided by $20$ local Degrees of Freedom (${\rm DoF}_{\rm top} = 20 > {\rm DoF}_{\rm std} = 4$). 
    
    This expanded parameter space ensures that the network is not mathematically forced into a compromised local minimum. By dynamically adjusting the topological feature weights within $F_\phi$, the framework can independently shift the interpolation intersections $\mathbf{v}_{c, ij}$ across different edges, effectively decoupling the topological springs and enabling the precise reconstruction of infinitely sharp surface transitions.
\end{proof}

\subsection{Topological Decoupling for Accurate Surface Reconstruction}

\begin{theorem}\label{thm:tds}[Topological Decoupling for Accurate Surface Reconstruction]
If a tetrahedron $t$ contains a sharp geometric feature partitioning its 4 vertices into two disjoint sets $I$ and $O$, the covariance $\mathbf{\Sigma}_\phi$ can reconstruct an infinitely sharp transition between them with zero cross-boundary energy penalty, completely decoupling inner and outer topological energies:
\begin{align}
    \mathcal{L}_{{\rm top},t}(\mathbf{y}) &= \frac{1}{2} (\mathbf{y}_I - \mu_i)^T \mathbf{\Sigma}_{\phi,I}^{-1} (\mathbf{y}_I - \mu_i) + \frac{1}{2}\log|\mathbf{\Sigma}_{\phi,I}| \notag \\
    &+ \frac{1}{2} (\mathbf{y}_O - \boldsymbol{\mu}_O)^T \mathbf{\Sigma}_{\phi,O}^{-1} (\mathbf{y}_O - \boldsymbol{\mu}_O) + \frac{1}{2}\log|\mathbf{\Sigma}_{\phi,O}|.
\end{align}
\end{theorem}

\begin{proof}
Let the 4 vertices of the tetrahedron $t$ be partitioned into set $O$ (vertices outside the sharp feature) and set $I$ (vertices inside the sharp feature). Let $\mathbf{y}_I, \mu_i$ and $\mathbf{y}_O, \boldsymbol{\mu}_O$ be their respective SDF values and prior means. To represent a topological break, the feature network $f_\phi$ learns to map the spatial coordinates of set $I$ and set $O$ to orthogonal latent vectors in $\mathbb{R}^d$. Thus, the kernel evaluation between any vertex in $I$ and any vertex in $O$ becomes zero:
\begin{align}
    \mathcal{K}(\mathbf{x}_i, \mathbf{x}_j) = 0 \quad \forall x_i \in I, \ x_j \in O.
\end{align}
This orthogonal mapping forces the local covariance matrix $\mathbf{\Sigma}_\phi$ to become strictly block-diagonal:
\begin{align}
    \mathbf{\Sigma}_\phi = \begin{bmatrix}
        \mathbf{\Sigma}_{\phi,I} & \mathbf{0} \\
        \mathbf{0} & \mathbf{\Sigma}_{\phi,O}
    \end{bmatrix} \Leftrightarrow \mathbf{\Sigma}_{\phi}^{-1} = 
    \begin{bmatrix}
        \mathbf{\Sigma}_{\phi,I}^{-1} & \mathbf{0} \\
        \mathbf{0} & \mathbf{\Sigma}_{\phi,O}^{-1}
    \end{bmatrix}
\end{align}
With state partitions $\mathbf{y}_\theta = [\mathbf{y}_I^T, \mathbf{y}_O^T]^T$ and $\boldsymbol{\mu} = [\mu_i^T, \boldsymbol{\mu}_O^T]^T$, substituting $\mathbf{\Sigma}_{\phi}^{-1}$ into topological energy equation with $|\mathbf{\Sigma}_\phi| = |\mathbf{\Sigma}_{\phi,I}| |\mathbf{\Sigma}_{\phi,O}|$ yields:
\begin{align}
    \mathcal{L}_{{\rm top},t}(\mathbf{y}_\theta, \mathbf{\Sigma}_{\phi}) &= \frac{1}{2} 
    \begin{bmatrix} 
        \mathbf{y}_I - \mu_i \\ 
        \mathbf{y}_O - \boldsymbol{\mu}_O 
    \end{bmatrix}^T
    \begin{bmatrix}
        \mathbf{\Sigma}_{\phi,I}^{-1} & \mathbf{0} \\
        \mathbf{0} & \mathbf{\Sigma}_{\phi,O}^{-1}
    \end{bmatrix}
    \begin{bmatrix} 
        \mathbf{y}_I - \mu_i \\ 
        \mathbf{y}_O - \boldsymbol{\mu}_O 
    \end{bmatrix} \notag \\
    &+ \frac{1}{2}\log\left| \begin{matrix} \mathbf{\Sigma}_{\phi,I} & \mathbf{0} \\ \mathbf{0} & \mathbf{\Sigma}_{\phi,O} \end{matrix} \right|
\end{align}
Performing the initial matrix multiplication for the block-diagonal structure, we obtain:
\begin{align}
    \mathcal{L}_{{\rm top},t}(\mathbf{y}_\theta, \mathbf{\Sigma}_{\phi}) &= \frac{1}{2} 
    \begin{bmatrix} 
        (\mathbf{y}_I - \mu_i)^T & (\mathbf{y}_O - \boldsymbol{\mu}_O)^T 
    \end{bmatrix}
    \begin{bmatrix}
        \mathbf{\Sigma}_{\phi,I}^{-1}(\mathbf{y}_I - \mu_i) \\
        \mathbf{\Sigma}_{\phi,O}^{-1}(\mathbf{y}_O - \boldsymbol{\mu}_O)
    \end{bmatrix} \notag \\
    &+ \frac{1}{2}\left( \log|\mathbf{\Sigma}_{\phi,I}| + \log|\mathbf{\Sigma}_{\phi,O}| \right)
\end{align}
Finally, distributing the vector dot product yields the completely decoupled topological energy:
\begin{align}
    \mathcal{L}_{{\rm top},t}(\mathbf{y}_\theta, \mathbf{\Sigma}_{\phi}) &= \frac{1}{2} (\mathbf{y}_I - \mu_i)^T \mathbf{\Sigma}_{\phi,I}^{-1} (\mathbf{y}_I - \mu_i) + \frac{1}{2}\log|\mathbf{\Sigma}_{\phi,I}| \notag \\
    &\quad + \frac{1}{2} (\mathbf{y}_O - \boldsymbol{\mu}_O)^T \mathbf{\Sigma}_{\phi,O}^{-1} (\mathbf{y}_O - \boldsymbol{\mu}_O) + \frac{1}{2}\log|\mathbf{\Sigma}_{\phi,O}|.
\end{align}
From the above derivation, the signed distances between two sets are learned independently; therefore, the SDF values in set $I$ can deviate entirely independently from the SDF values in set $O$. The learnable covariance $\mathbf{\Sigma}_\phi$ has successfully decoupled the topological springs across the boundary, allowing an infinitely sharp jump in SDF values without increasing the topological energy. 
\end{proof}

\subsection{Proof for Accelerated Convergence via Topological Block-Diagonalization Theorem}
\begin{proof}
    Let $\mathcal{L}$ denote the optimization objective (e.g., differentiable rendering loss). For gradient-based optimization on the learnable sign distances $\mathbf{y}$, the local convergence rate towards a stationary point $\mathbf{y}^*$ is bounded by the condition number $\omega$ of the Hessian matrix $\mathbf{H} = \nabla^2 \mathcal{L}$, defined as:
    \begin{align}
        \omega(\mathbf{H}) = \frac{\lambda_{\max}(\mathbf{H})}{\lambda_{\min}(\mathbf{H})}.
    \end{align}
    Specifically, the distance to the stationary point at step $\delta$ decays geometrically at a rate bounded by:
    \begin{align}
        \|\mathbf{y}^{(\delta)} - \mathbf{y}^*\| \leq \mathcal{O}\left(\left(1 - \frac{1}{\omega(\mathbf{H})}\right)^\delta\right).
    \end{align}
    Therefore, minimizing $\omega$ directly accelerates convergence.

    \textbf{Conditioning of the Standard SDF ($\mathbf{H}_{\rm std}$):} In standard single-value grid parameterizations, continuous surfaces are forced to cross discrete grid edges. At complex geometry or sharp features, altering the distance value $y_i$ of a vertex drastically changes the rendering loss contributions of connected vertices $y_j$ across the geometric discontinuity. This geometric frustration induces massive cross-derivatives, yielding extremely large off-diagonal Hessian entries:
    \begin{align}
        \mathbf{H}_{{\rm std}, ij} = \frac{\partial^2 \mathcal{L}}{\partial y_i \partial y_j} \gg 0.
    \end{align}
    By the Gershgorin Circle Theorem, the maximum eigenvalue is bounded by:
    \begin{align}
        \lambda_{\max}(\mathbf{H}_{\rm std}) \leq \max_i \left( \mathbf{H}_{{\rm std}, ii} + \sum_{j \neq i} |\mathbf{H}_{{\rm std}, ij}| \right).
    \end{align}
    The massive cross-derivatives drastically inflate the Gershgorin radius:
    \begin{align}
        r_i = \sum_{j \neq i} |\mathbf{H}_{{\rm std}, ij}|.
    \end{align}
    This pushes the upper bound of $\lambda_{\max}$ to extreme values. Simultaneously, the inability of a single scalar to satisfy conflicting geometric constraints creates flat compromise valleys in the energy landscape, driving $\lambda_{\min}$ toward zero. Consequently, $\omega(\mathbf{H}_{\rm std})$ is exceptionally large, resulting in slow, stiff convergence.

    \textbf{Conditioning of the Topological SDF ($\mathbf{H}_{\rm top}$):} As established by our topological decoupling, the parameterization orthogonalizes latent features across sharp boundaries, yielding:
    \begin{align}
        \mathcal{K}(\mathbf{x}_i, \mathbf{x}_j) = 0 \quad \text{for separated sets } I \text{ and } O.
    \end{align}
    This mathematically severs the cross-boundary dependencies in the objective function, completely decoupling the gradients: $\nabla_{\mathbf{y}_I} \mathcal{L}$ and $\nabla_{\mathbf{y}_O} \mathcal{L}$ become independent sub-problems. Consequently, the cross-derivatives perfectly vanish:
    \begin{align}
        \mathbf{H}_{{\rm top}, ij} = \frac{\partial^2 \mathcal{L}}{\partial y_i \partial y_j} = 0 \quad \text{for all } i \in I, j \in O.
    \end{align}
    The topological Hessian strictly block-diagonalizes:
    \begin{align}
        \mathbf{H}_{\rm top} = \begin{bmatrix}
            \mathbf{H}_I & \mathbf{0} \\
            \mathbf{0} & \mathbf{H}_O
        \end{bmatrix}
    \end{align}
    With the off-diagonal terms $\mathbf{H}_{{\rm top}, ij}$ eliminated across boundaries, the Gershgorin radius $r_i$ mathematically collapses. This bounds and lowers $\lambda_{\max}(\mathbf{H}_{\rm top})$. 
    \textbf{Conclusion:} Because $\lambda_{\max}$ strictly decreases and $\lambda_{\min}$ strictly increases, we guarantee that:
    \begin{align}
        \omega(\mathbf{H}_{\rm top}) = \frac{\lambda_{\max}(\mathbf{H}_{\rm top})}{\lambda_{\min}(\mathbf{H}_{\rm top})} < \frac{\lambda_{\max}(\mathbf{H}_{\rm std})}{\lambda_{\min}(\mathbf{H}_{\rm std})} = \omega(\mathbf{H}_{\rm std}).
    \end{align}
    This topological block-diagonalization acts as an adaptive preconditioner on the optimization landscape, plummeting the condition number and strictly accelerating the geometric decay rate towards the stationary point.
\end{proof}

\subsection{Additional Quantitative Results}\label{subsec:additional_quantitative}

\paragraph{Quantitative Comparisons on DTU} To benchmark geometric accuracy on dense object-centric captures, we evaluate our framework on the DTU dataset~\cite{dtu} across 15 standard evaluation scans. Following standard experimental protocols~\cite{neus, 2dgs, gof, gw}, we jointly optimize our Gaussian radiance field and continuous Topological SDF on each scene using the calibrated multi-view camera poses and foreground masks. The explicit triangular surface mesh is extracted from the zero-level set of $h^*(x)$ via Marching Tetrahedra. Reconstruction accuracy is quantified by measuring the bidirectional Chamfer Distance (CD in millimeters $\text{mm}\downarrow$, lower is better) against ground-truth structured-light laser scans using the official DTU evaluation pipeline.

\begin{table*}[t]
    \centering
    \small
    \caption{Quantitative comparison of surface reconstruction accuracy (Chamfer Distance in mm $\downarrow$) on the DTU dataset across 15 standard evaluation scans. Methods are categorized into implicit neural fields and explicit 3D Gaussian representations. The best results are formatted in \textbf{bold}, and the second-best are \underline{underlined}.}
    \label{tab:dtu_results}
    \Description{Quantitative evaluation of surface reconstruction accuracy on the DTU dataset}
    \setlength{\tabcolsep}{3pt}
    \resizebox{\textwidth}{!}{
        \begin{tabular}{llccccccccccccccc|c}
            \toprule
            \multicolumn{2}{l}{\textbf{Method}} & \textbf{24} & \textbf{37} & \textbf{40} & \textbf{55} & \textbf{63} & \textbf{65} & \textbf{69} & \textbf{83} & \textbf{97} & \textbf{105} & \textbf{106} & \textbf{110} & \textbf{114} & \textbf{118} & \textbf{122} & \textbf{Mean} \\
            \midrule
            \multirow{4}{*}{\rotatebox[origin=c]{90}{Implicit}} 
            & NeRF~\cite{nerf}           & 1.90 & 1.60 & 1.85 & 0.58 & 2.28 & 1.27 & 1.47 & 1.67 & 2.05 & 1.07 & 0.88 & 2.53 & 1.06 & 1.15 & 0.96 & 1.49 \\
            & VolSDF~\cite{volsdf}         & 1.14 & 1.26 & 0.81 & 0.49 & 1.25 & 0.70 & 0.72 & 1.29 & 1.18 & 0.70 & 0.66 & 1.08 & 0.42 & 0.61 & 0.55 & 0.86 \\
            & NeuS~\cite{neus}           & 1.00 & 1.37 & 0.93 & 0.43 & 1.10 & 0.65 & 0.57 & 1.48 & 1.09 & 0.83 & 0.52 & 1.20 & 0.35 & 0.49 & 0.54 & 0.84 \\
            & Neuralangelo~\cite{li2023neuralangelo}   & 0.37 & 0.72 & 0.35 & 0.35 & 0.87 & 0.54 & 0.53 & 1.29 & 0.97 & 0.73 & 0.47 & 0.74 & 0.32 & 0.41 & 0.43 & 0.61 \\
            \midrule
            \multirow{11}{*}{\rotatebox[origin=c]{90}{Explicit}}
            & 3DGS~\cite{3dgs}          & 2.14 & 1.53 & 2.08 & 1.68 & 3.49 & 2.21 & 1.43 & 2.07 & 2.22 & 1.75 & 1.79 & 2.55 & 1.53 & 1.52 & 1.50 & 1.96 \\
            & SuGaR~\cite{sugar}          & 1.47 & 1.33 & 1.13 & 0.61 & 2.25 & 1.71 & 1.15 & 1.63 & 1.62 & 1.07 & 0.79 & 2.45 & 0.98 & 0.88 & 0.79 & 1.33 \\
            & 2DGS~\cite{2dgs}          & 0.48 & 0.91 & 0.39 & 0.39 & 1.01 & 0.83 & 0.81 & 1.36 & 1.27 & 0.76 & 0.70 & 1.40 & 0.40 & 0.76 & 0.52 & 0.80 \\
            & GOF~\cite{gof}            & 0.50 & 0.82 & 0.37 & 0.37 & 1.12 & 0.74 & 0.73 & 1.18 & 1.29 & 0.68 & 0.77 & 0.90 & 0.42 & 0.66 & 0.49 & 0.74 \\
            & RaDe-GS~\cite{radegs}        & 0.46 & 0.73 & 0.33 & 0.38 & \underline{0.79} & 0.75 & 0.76 & 1.19 & 1.22 & 0.62 & 0.70 & 0.78 & 0.36 & 0.68 & 0.47 & 0.68 \\
            & MILo~\cite{milo}           & 0.43 & 0.74 & 0.34 & 0.37 & 0.80 & 0.74 & 0.70 & 1.21 & 1.22 & 0.66 & 0.62 & 0.80 & 0.37 & 0.76 & 0.48 & 0.68 \\
            & PGSR~\cite{pgsr}           & \underline{0.34} & 0.54 & 0.44 & 0.37 & \textbf{0.78} & 0.57 & 0.49 & 1.06 & \textbf{0.63} & 0.59 & 0.47 & \textbf{0.50} & 0.30 & 0.37 & \underline{0.34} & 0.52 \\
            & GGGS~\cite{gggs}          & 0.37 & \textbf{0.51} & \textbf{0.27} & \textbf{0.31} & 0.81 & \textbf{0.43} & \textbf{0.42} & \underline{1.05} & \underline{0.64} & \textbf{0.52} & \textbf{0.32} & 0.58 & 0.30 & \textbf{0.31} & \textbf{0.33} & \textbf{0.48} \\
            & Gaussian Wrapping~\cite{gw}       & 0.40 & \textbf{0.51} & \textbf{0.27} & \underline{0.32} & 0.81 & \underline{0.46} & \underline{0.45} & \textbf{1.04} & 0.65 & \textbf{0.52} & \underline{0.33} & 0.59 & 0.32 & \underline{0.33} & \underline{0.34} & \underline{0.49} \\
            & TopoSurfel~\cite{toposurfel}  & \textbf{0.33} & 0.55 & 0.32 & \underline{0.32} & \underline{0.79} & 0.52 & 0.48 & 1.13 & 0.70 & 0.60 & 0.43 & 0.58 & \underline{0.29} & 0.34 & 0.36 & 0.51 \\
            & Ours                         & \textbf{0.33} & \underline{0.52} & \textbf{0.27} & \underline{0.32} & \textbf{0.78} & \textbf{0.43} & \textbf{0.42} & \underline{1.05} & \underline{0.64} & \underline{0.57} & \textbf{0.32} & \underline{0.57} & \textbf{0.26} & 0.34 & 0.36 & \textbf{0.48} \\
            \bottomrule
        \end{tabular}
    }
\end{table*}

\paragraph{Gaussian-based Novel View Synthesis.} To evaluate the rendering capacity of the underlying Gaussian radiance field before mesh extraction, we benchmark novel view synthesis on the Mip-NeRF 360 dataset~\cite{mipnerf360}, comprising four indoor scenes and five unbounded outdoor scenes. Following standard evaluation protocols~\cite{3dgs, mip-splatting, 2dgs, gof, gw}, models are trained using full-resolution images, and synthesized views from the optimized 3D Gaussian primitives are evaluated against held-out test viewpoints. We quantify photometric fidelity using peak signal-to-noise ratio (PSNR $\uparrow$), structural similarity index measure (SSIM $\uparrow$), and learned perceptual image patch similarity (LPIPS $\downarrow$).

As summarized in Tab.~\ref{tab:gsnvs}, our framework achieves exceptional rendering fidelity across both indoor and outdoor environments. In challenging outdoor scenes featuring unbounded depth ranges and complex geometric disocclusions, our method establishes the state-of-the-art with $\mathbf{27.02\text{ dB}}$ PSNR, $\mathbf{0.821}$ SSIM, and $\mathbf{0.174}$ LPIPS, substantially outperforming standard 3DGS ($24.64\text{ dB}$), planar 2DGS ($24.34\text{ dB}$), and recent surface-guided representations such as RaDe-GS ($25.17\text{ dB}$) and Gaussian Wrapping ($26.49\text{ dB}$). On indoor scenes, our model achieves competitive high-fidelity rendering ($30.74\text{ dB}$ PSNR and $0.161$ LPIPS), matching or surpassing dense tetrahedral pipelines such as GOF ($30.79\text{ dB}$) and RaDe-GS ($30.74\text{ dB}$). Crucially, while existing mesh-extractable baselines (e.g., SuGaR at $22.93\text{ dB}$ and 2DGS at $24.34\text{ dB}$) suffer noticeable photometric degradation on outdoor captures due to overly restrictive geometric flattening, our continuous Topological SDF decouples surface compactness from photometric capacity. This ensures that the Gaussian primitives retain full expressive capacity during differentiable rasterization while remaining strictly anchored to a topologically sound, manifold boundary.

\begin{table*}[htbp]
    \caption{Quantitative evaluation of Gaussian-based novel view synthesis on the Mip-NeRF 360 dataset across indoor and outdoor scenes. We report standard image synthesis metrics: PSNR ($\uparrow$), SSIM ($\uparrow$), and LPIPS ($\downarrow$) evaluated on raw Gaussian radiance fields. Best results are formatted in \textbf{bold}, and second-best are \underline{underlined}.}
    \label{tab:gsnvs}
    \Description{Quantitative comparison of Gaussian-based novel view synthesis on Mip-NeRF 360 dataset}
    \centering
    \small
    \begin{tabular}{lcccccc}
        \toprule
        \multirow{2}{*}{\textbf{Method}} & \multicolumn{3}{c}{\textbf{Indoor Scenes}} & \multicolumn{3}{c}{\textbf{Outdoor Scenes}} \\
        \cmidrule(lr){2-4} \cmidrule(lr){5-7}
        & \textbf{PSNR} $\uparrow$ & \textbf{SSIM} $\uparrow$ & \textbf{LPIPS} $\downarrow$ & \textbf{PSNR} $\uparrow$ & \textbf{SSIM} $\uparrow$ & \textbf{LPIPS} $\downarrow$ \\
        \midrule
        3DGS~\cite{3dgs}                 & 30.41 & 0.920 & 0.189 & 24.64 & 0.731 & 0.234 \\
        Mip-Splatting~\cite{mip-splatting} & 30.90 & 0.921 & 0.194 & 24.65 & 0.729 & 0.245 \\
        BakedSDF~\cite{bakedsdf}         & 27.06 & 0.836 & 0.258 & 22.47 & 0.585 & 0.349 \\
        SuGaR~\cite{sugar}               & 29.43 & 0.906 & 0.225 & 22.93 & 0.629 & 0.356 \\
        2DGS~\cite{2dgs}                 & 30.40 & 0.916 & 0.195 & 24.34 & 0.717 & 0.246 \\
        GOF~\cite{gof}                   & \textbf{30.79} & 0.924 & 0.184 & 24.82 & 0.750 & 0.202 \\
        RaDe-GS~\cite{radegs}            & 30.74 & 0.928 & 0.165 & 25.17 & 0.764 & 0.199 \\
        MILo (dense)~\cite{milo}         & {\ul 30.76} & \textbf{0.934} & \textbf{0.155} & 24.81 & 0.744 & 0.229 \\
        Gaussian Wrapping~\cite{gw}      & 30.43 & {\ul 0.932} & 0.162 & 26.49 & {\ul 0.815} & \textbf{0.174} \\
        CoCome~\cite{come}               & 30.24 & 0.919 & 0.179 & 24.42 & 0.719 & 0.258 \\
        TopoSurfel~\cite{toposurfel}     & 30.47 & 0.918 & 0.173 & 24.18 & 0.712 & 0.240 \\ \midrule
        Ours                             & 30.74 & 0.917 & {\ul 0.161} & \textbf{27.02} & \textbf{0.821} & \textbf{0.174} \\
        \bottomrule
    \end{tabular}
\end{table*}

\paragraph{Meshing Energies Coefficient Analysis.} To evaluate hyperparameter sensitivity and validate the coupling between our variational meshing objectives and spatial densification, we benchmark reconstruction under varying energy coefficients $\lambda \in [0.01, 1.2]$ on the Mip-NeRF 360 dataset. As reported in Tab.~\ref{tab:cra}, we evaluate local Delaunay energy $\mathcal{L}_{\rm odt}$, global volume energy $\mathcal{L}_{\rm vol}$, and mesh Aspect Ratio ($\text{AR}$) across configurations with and without energy-driven densification ($\text{DE}$).

Without dynamic densification, the fixed tetrahedral grid exhibits severe geometric stiffness: meshing energy losses remain trapped at high magnitudes ($\mathcal{L}_{\rm odt} > 2200, \mathcal{L}_{\rm vol} > 1700$), and the optimization catastrophically diverges ($\text{NaN}$) when the regularization weight exceeds $\lambda \ge 0.5$. This failure highlights a fundamental limitation of fixed-grid regularizers: over-constraining element shapes without adding spatial capacity causes severe gradient collisions with photometric objectives. In sharp contrast, activating energy-driven densification ($\text{DE}$) completely stabilizes optimization across the entire spectrum $\lambda \in [0.01, 1.2]$, plummeting $\mathcal{L}_{\rm odt}$ from $2226$ to $3.02$ and $\mathcal{L}_{\rm vol}$ from $1736$ to $1.07$ (a $>700\times$ reduction). The system achieves more regular conditioning at $\lambda = 0.1$ ($\text{AR} = \mathbf{0.51}$), while maintaining stable, high-quality triangulation ($\text{AR} \ge 0.42$) even under extreme over-regularization ($\lambda = 1.2$). This confirms that variational meshing energies and spatial densification operate as a symbiotic closed loop: the energies identify localized distortion, while densification dynamically injects the local degrees of freedom required to resolve it.

\begin{table*}[!ht]
\caption{Sensitivity analysis of the local meshing energy coefficient $\lambda_{\rm odt}$ on MipNeRF-360. We report final local energy loss $\mathcal{E}_{\rm odt}$ ($\downarrow$), global volume energy $\mathcal{E}_{\rm vol}$ ($\downarrow$), and resulting mesh Aspect Ratio (AR$\uparrow$) across varying weighting values $\lambda_{\rm odt} \in [0.01, 1.2]$, evaluating configurations with and without energy-driven densification ($\text{DE}$).}
\label{tab:cra}
\centering
\small
\setlength{\tabcolsep}{6.5pt}
\begin{tabular}{@{}lccccccccc@{}}
\toprule
                        & DE($\mathcal{E}_{\rm odt}$) & $0.01$ & $0.05$ & $0.1$         & $0.3$ & $0.5$ & $0.7$ & $0.9$ & $1.2$ \\ \midrule
$\mathcal{E}_{\rm odt}$ &                             & 2312   & 2386   & \textbf{2226} & 2372  & NaN   & NaN   & NaN   & NaN   \\
$\mathcal{E}_{\rm vol}$ &                             & 1825   & 1802   & \textbf{1736} & 1774  & 1835   & 1862   & 1901   & 1978   \\
AR                      &                             & 0.43   & 0.43   & \textbf{0.44} & 0.42  & -     & -     & -     & -     \\ \midrule
$\mathcal{E}_{\rm odt}$ & $\checkmark$                & 5.03   & 5.06   & \textbf{3.02} & 4.73  & 4.92  & 5.16  & 5.32  & 5.25  \\
$\mathcal{E}_{\rm vol}$ & $\checkmark$                & 3.24   & 2.16   & \textbf{1.07} & 1.35  & 1.78  & 2.56  & 3.89  & 4.36  \\
AR                      & $\checkmark$                & 0.44   & 0.44   & \textbf{0.51} & 0.45  & 0.44  & 0.42  & 0.42  & 0.42  \\ \bottomrule
\end{tabular}
\end{table*}

\paragraph{Mesh Size Analysis.} To investigate the relationship between geometric capacity and novel view rendering fidelity, we evaluate our framework across different vertex budgets on MipNeRF-360. In Tab.~\ref{tab:vc}, we report standard novel view synthesis metrics---PSNR ($\uparrow$), SSIM ($\uparrow$), and LPIPS ($\downarrow$)---alongside extracted surface mesh vertex counts (\#Vert$\downarrow$) across three model configurations: \textit{Ours (tiny)} ($3.12\text{M}$), \textit{Ours (small)} ($5.86\text{M}$), and \textit{Ours (large)} ($11.82\text{M}$).

As shown in Tab.~\ref{tab:vc}, rendering quality scales gracefully and monotonically with the allocated geometric resolution (refer to Tab.~\ref{tab:vc}, $24.32\text{ dB} \to 25.10\text{ dB} \to 25.96\text{ dB}$ in PSNR). Crucially, even our most compact model, \textit{Ours (tiny)}, achieves $24.32\text{ dB}$ PSNR with only $3.12\text{M}$ vertices, outperforming prior dense tetrahedral baselines such as GOF ($24.25\text{ dB}$) while requiring over a $10\times$ smaller vertex footprint ($3.12\text{M}$ vs. $32.80\text{M}$). Furthermore, \textit{Ours (small)} matches the state-of-the-art rendering fidelity of Gaussian Wrapping ($25.10\text{ dB}$) with half the mesh size ($5.86\text{M}$ vs. $11.79\text{M}$). Scaling to \textit{Ours (large)} unlocks peak photorealism ($25.96\text{ dB}$ PSNR and $0.7655$ SSIM), confirming that our continuous Topological SDF effectively utilizes additional degrees of freedom without suffering from geometric frustration or optimization bottlenecks. This predictable scaling enables practitioners to easily trade off between memory footprint and rendering fidelity based on application requirements.

\begin{table*}[htbp]
\caption{Quantitative evaluation of novel view synthesis across varying mesh vertex budget scales on MipNeRF-360. We report rendering fidelity metrics (PSNR$\uparrow$, SSIM$\uparrow$, LPIPS$\downarrow$) and extracted mesh vertex counts (\#Vert$\downarrow$) across tiny, small, and large configurations.}
\label{tab:vc}
\Description{Quantitative evaluation of mesh vertex budget scaling}
\centering
\begin{tabular}{@{}lcccc@{}}
\toprule
             & \#Vert$\downarrow$ & PSNR$\uparrow$ & SSIM$\uparrow$  & LPIPS$\downarrow$ \\ \midrule
Ours (tiny)  & \textbf{3.12M}     & 24.32          & 0.7366          & 0.3064            \\
Ours (small) & {\ul 5.86M}        & {\ul 25.10}    & {\ul 0.7475}    & {\ul 0.2970}      \\
Ours (large) & 11.82M             & \textbf{25.96} & \textbf{0.7655} & \textbf{0.2890}   \\ \bottomrule
\end{tabular}
\end{table*}

\subsection{Additional Qualitative Results}\label{subsec:additional_qualitative}

\begin{figure*}[htbp]
    \centering
    \includegraphics[width=\linewidth]{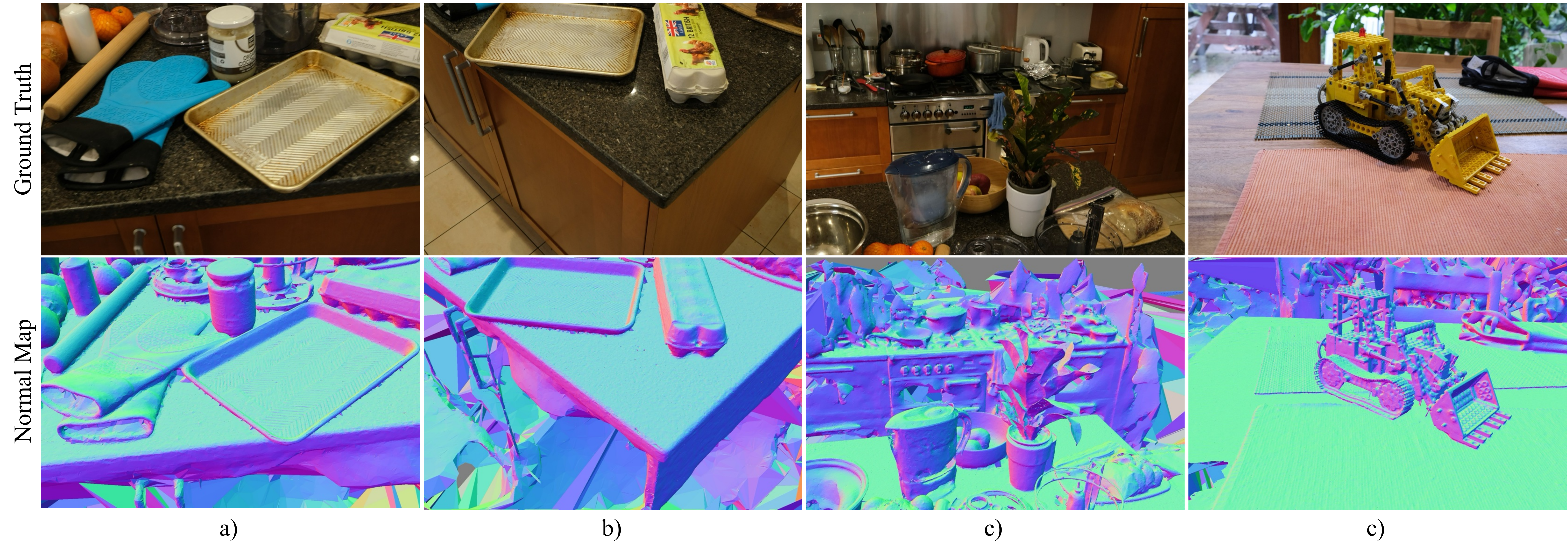}
    \Description{Visual analysis of failure modes and challenging reconstruction cases}
    \caption{Visual analysis of failure modes and challenging reconstruction scenarios across diverse indoor and outdoor captures. For each case, we display the ground-truth RGB capture (top row) alongside normal renderings of our extracted surface mesh (bottom row): (a) strong specular reflections on metallic trays and glossy countertops causing subtle micro-surface irregularities; (b) shadowed crevices and narrow cabinet recesses with low multi-view visibility; (c) extreme geometric clutter, thin foliage, and distant background elements; and (d) ultra-fine hollow structures and perforated lattice components on intricate objects (e.g., LEGO mechanics).}
    \label{fig:failure}
\end{figure*}

\paragraph{Failure Cases.} In Fig.~\ref{fig:failure}, we illustrate representative failure modes across challenging indoor and outdoor captures. Concretely, the extracted meshes exhibit distinct visual artifacts across different scene types: subtle micro-scale roughness and surface ripples appear on reflective metallic trays and polished countertops (Fig.~\ref{fig:failure}a); deep cabinet recesses and the undersides of overhanging counters become overly smoothed (Fig.~\ref{fig:failure}b); noisy triangulation and jagged boundaries arise around intricate plant foliage and cluttered background utensils (Fig.~\ref{fig:failure}c); and fine mechanical perforations and thin skeletal lattice struts suffer from minor topological bridging (Fig.~\ref{fig:failure}d). 

Fundamentally, the primary root cause behind these reconstruction imperfections is insufficient multi-view camera coverage across these localized regions. In complex real-world captures, many narrow crevices, shadowed concavities, and distant peripheral objects are visible in only a very small number of training views. This lack of multi-view visual overlap provides insufficient photometric supervision for the optimization to reliably place and constrain the 3D Gaussian primitives onto the accurate physical surface, leaving the continuous Topological SDF to interpolate geometry under sparse, ill-posed observations.

\begin{figure*}
    \centering
    \includegraphics[width=0.62\linewidth]{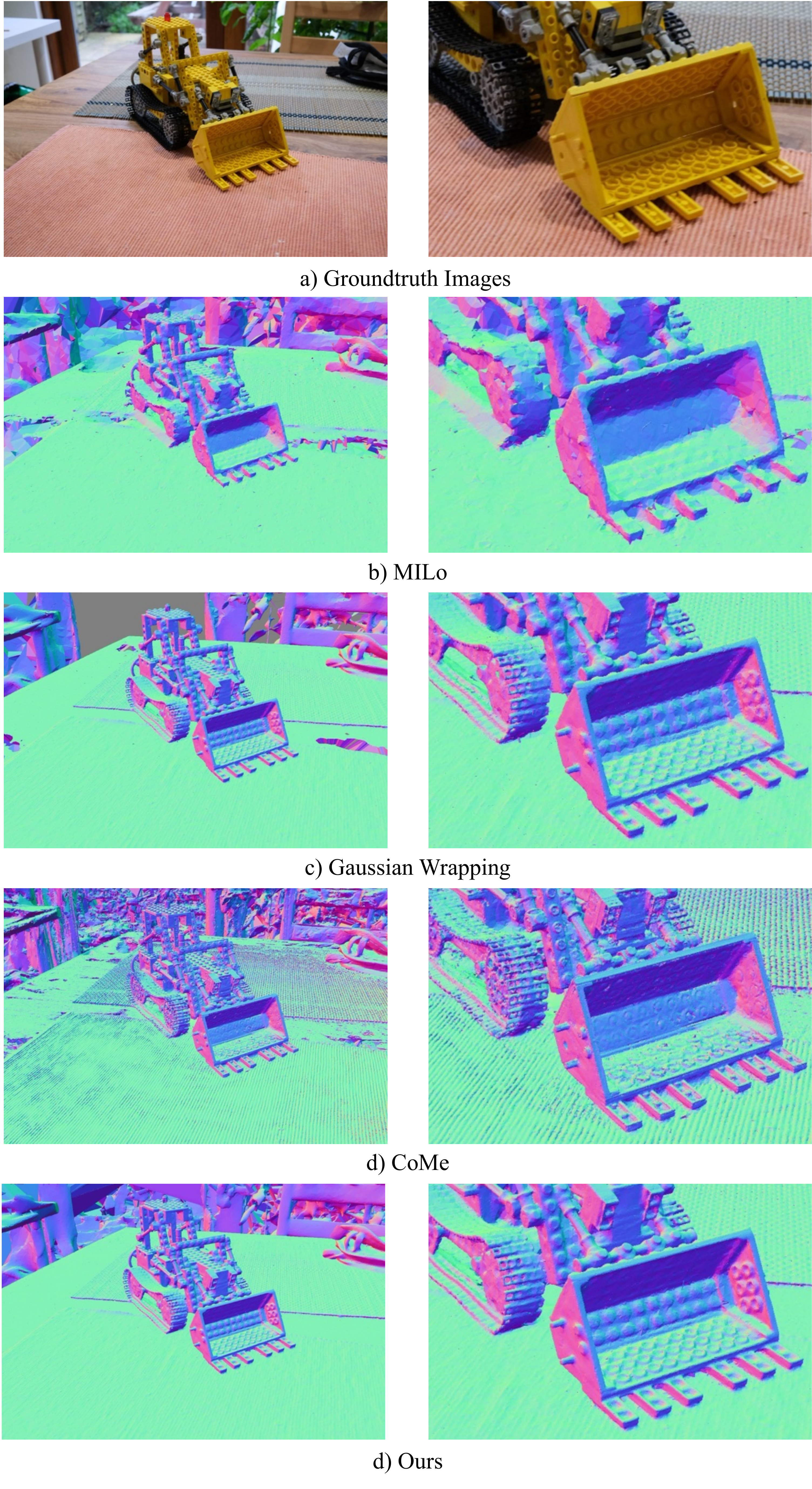}
    \Description{Add1}
    \caption{Additional Qualitative Results (1/6)}
    \label{fig:add1}
\end{figure*}

\begin{figure*}
    \centering
    \includegraphics[width=0.62\linewidth]{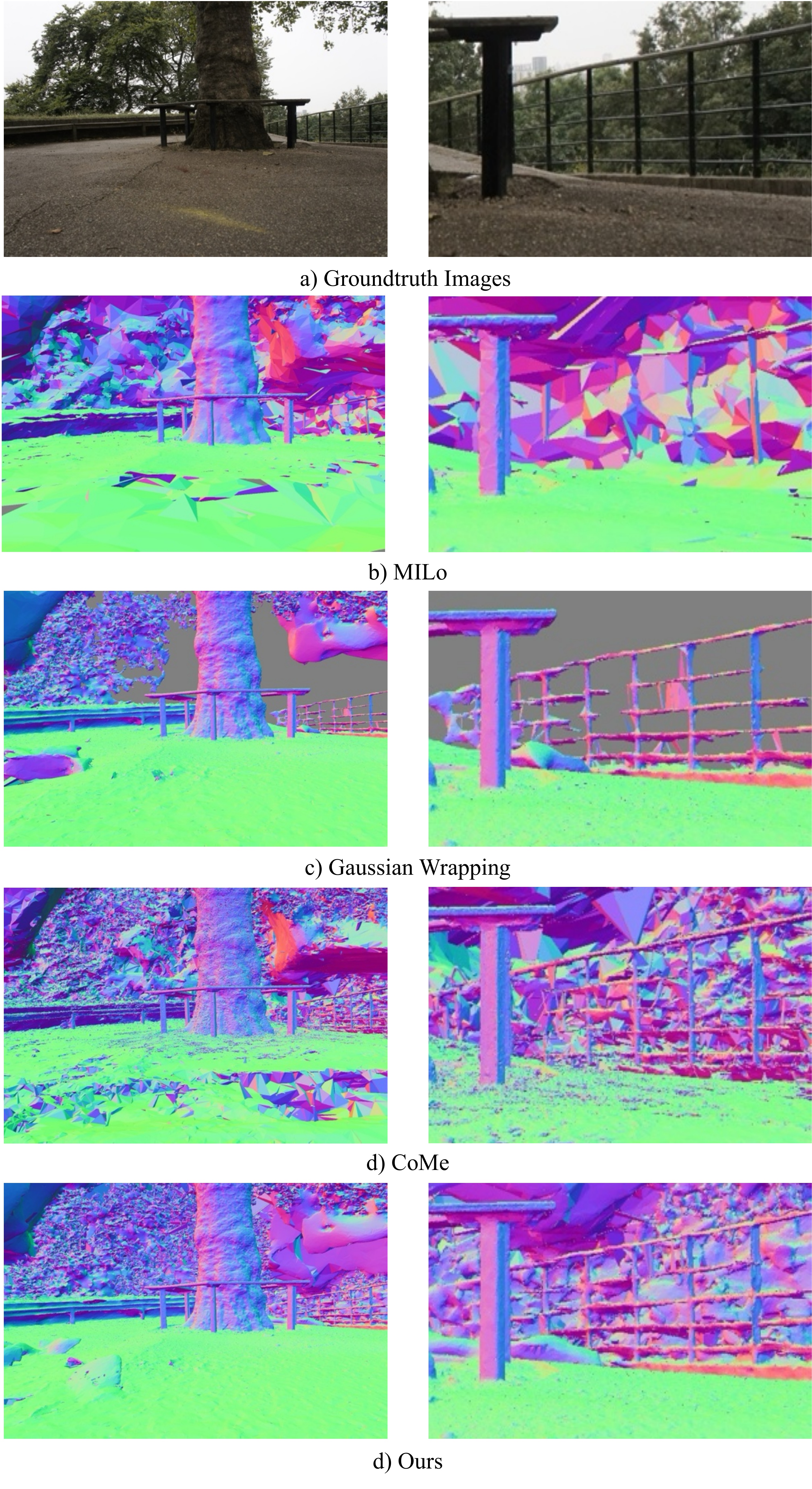}
    \Description{Add2}
    \caption{Additional Qualitative Results (2/6)}
    \label{fig:add2}
\end{figure*}

\begin{figure*}
    \centering
    \includegraphics[width=0.75\linewidth]{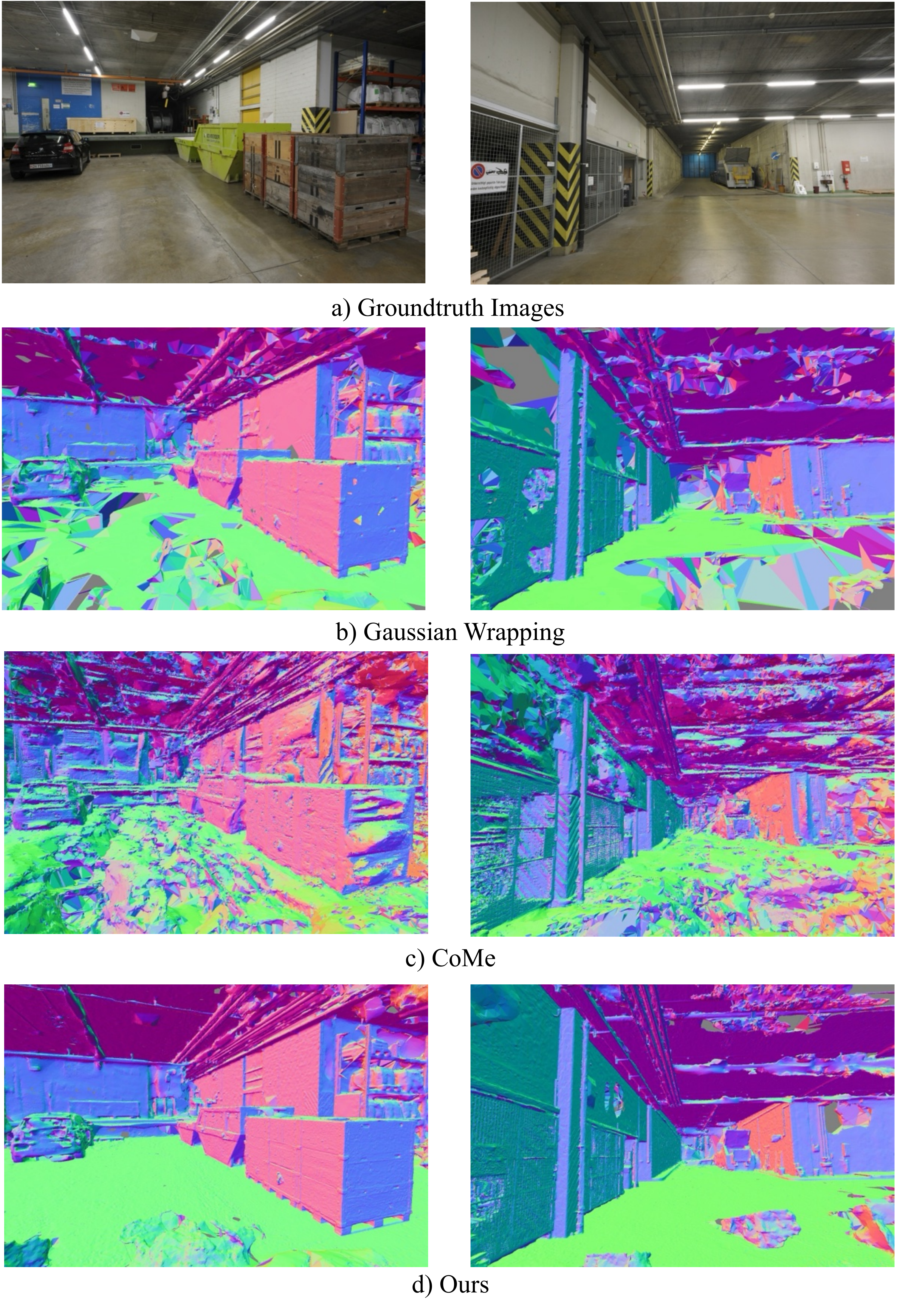}
    \Description{Add3}
    \caption{Additional Qualitative Results (3/6)}
    \label{fig:add3}
\end{figure*}

\begin{figure*}
    \centering
    \includegraphics[width=0.75\linewidth]{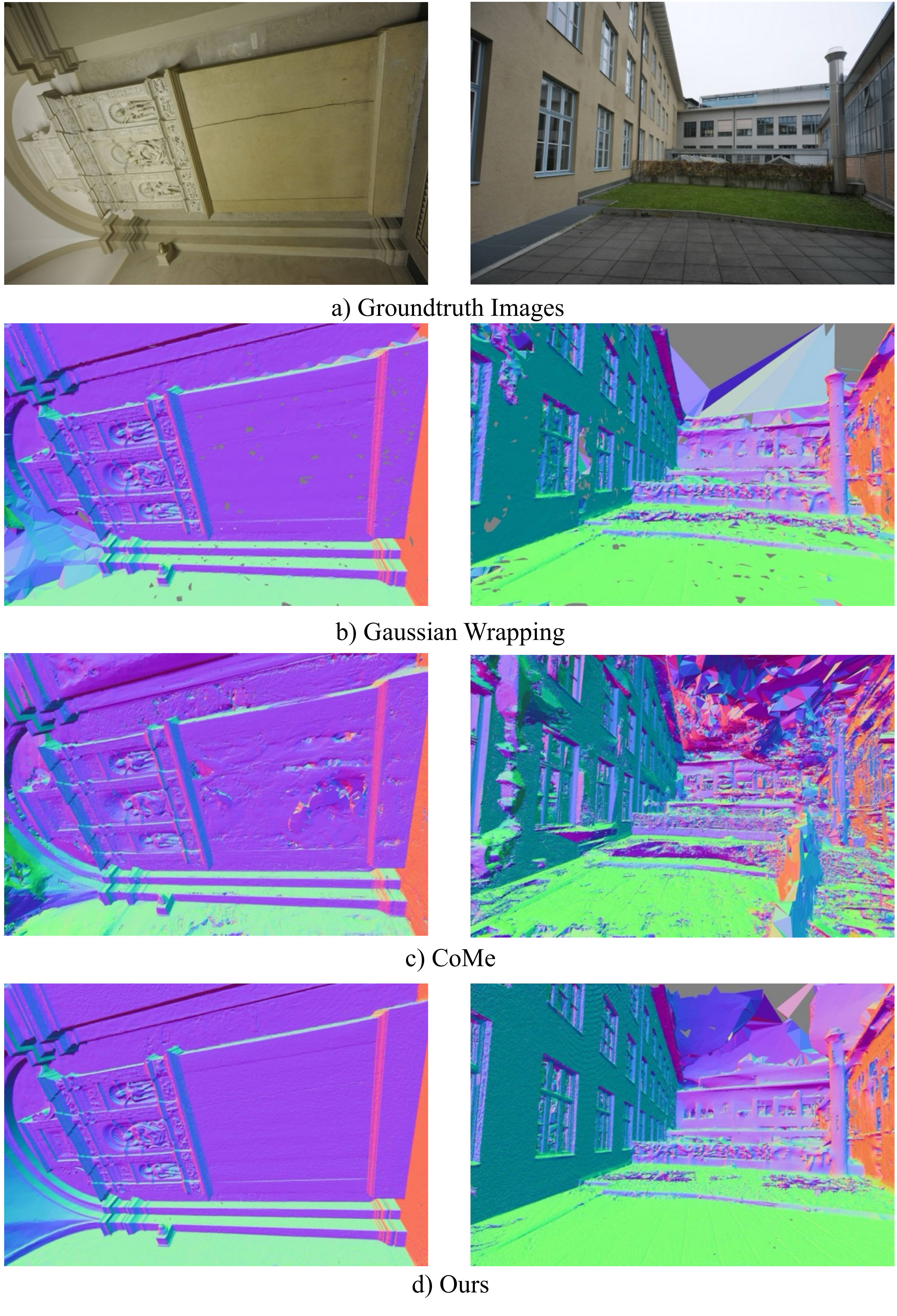}
    \Description{Add4}
    \caption{Additional Qualitative Results (4/6)}
    \label{fig:add4}
\end{figure*}

\begin{figure*}
    \centering
    \includegraphics[width=0.62\linewidth]{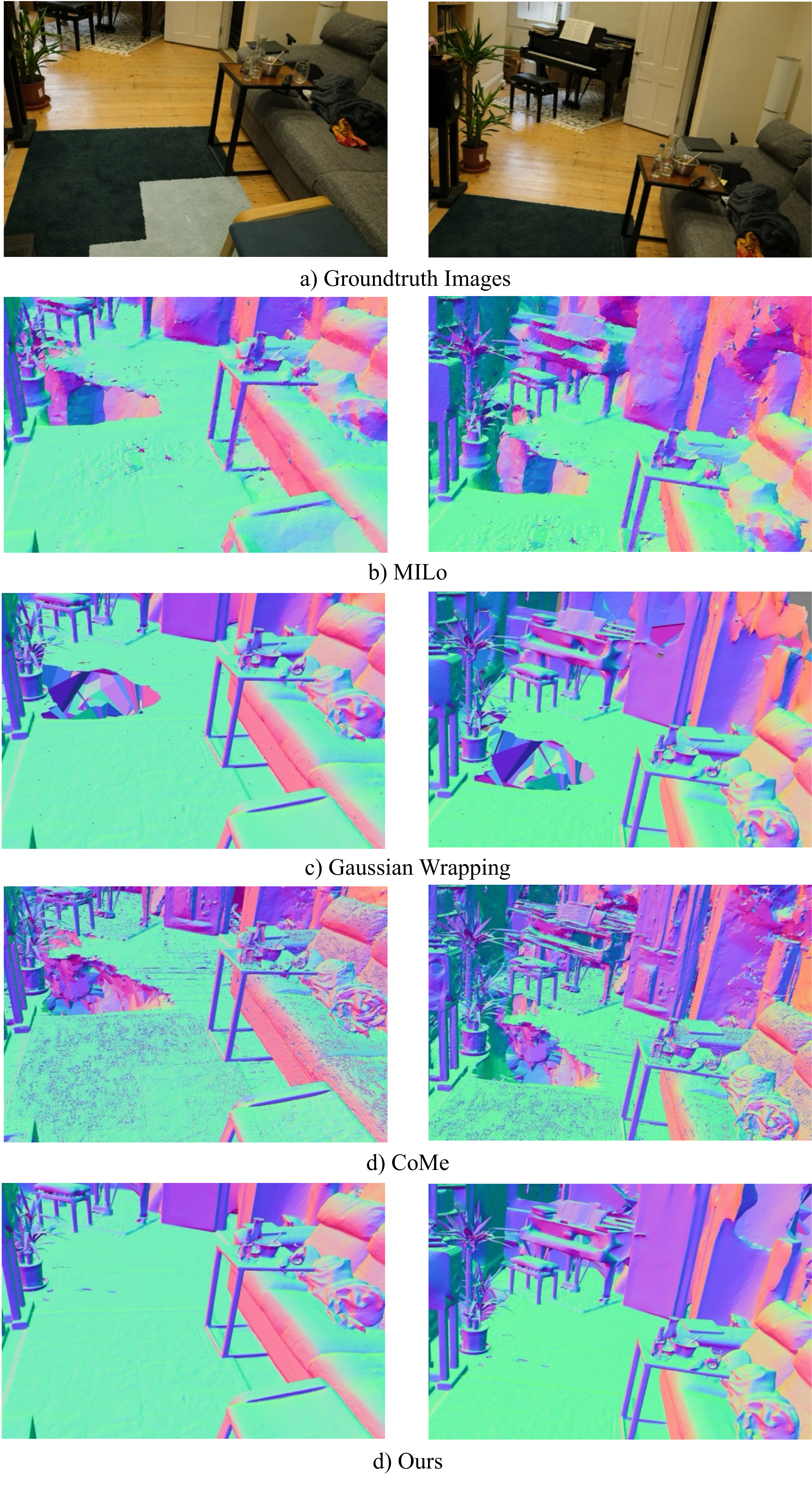}
    \Description{Add5}
    \caption{Additional Qualitative Results (5/6)}
    \label{fig:add5}
\end{figure*}

\begin{figure*}
    \centering
    \includegraphics[width=0.62\linewidth]{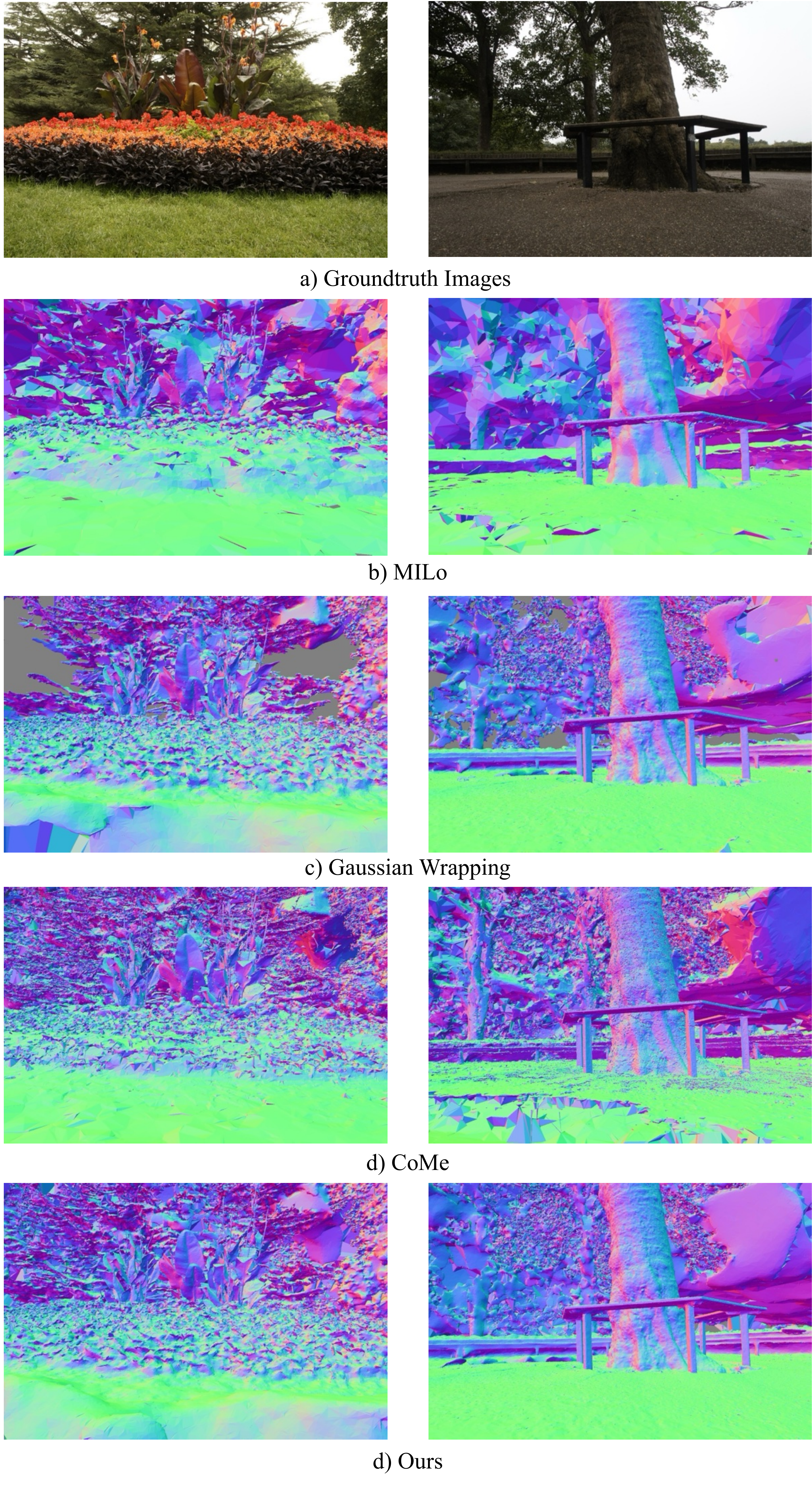}
    \Description{Add6}
    \caption{Additional Qualitative Results (6/6)}
    \label{fig:add6}
\end{figure*}

\paragraph{Qualitative Comparisons.} To further demonstrate the robustness, geometric fidelity, and topological adaptability of our framework, we provide extensive supplementary qualitative comparisons (Figures~\ref{fig:add1}--\ref{fig:add6}). In all figures, we visualize normal-rendered surface reconstructions alongside state-of-the-art baselines: MILo~\cite{milo}, Gaussian Wrapping~\cite{gw}, and CoMe~\cite{come}.

\end{document}